\documentclass{article}
\usepackage{graphicx} % Required for inserting images
\usepackage{url} %this package should fix any errors with URLs in refs.
\usepackage{lineno}
\usepackage[inline]{trackchanges}
\usepackage{soul}
\usepackage{amsmath}
\usepackage{tabularx}
\usepackage{authblk}
\usepackage{bm}
\usepackage[export]{adjustbox}
\usepackage[margin=2cm]{geometry}

\title{Ensemble weather forecast post-processing with a flexible probabilistic neural network approach}
\author[1,2]{Peter Mlakar}
\author[2]{Janko Merše}
\author[1]{Jana Faganeli Pucer}

\affil[1]{University of Ljubljana, Faculty of Computer and Information Science, Ljubljana}
\affil[2]{Slovenian Environment Agency, Slovenia}

\begin{document}
\maketitle

\section*{Abstract}
Ensemble forecast post-processing is a necessary step in producing accurate probabilistic forecasts.
Conventional post-processing methods operate by estimating the parameters of a parametric distribution, frequently on a per-location or per-lead-time basis, which limits their expressive power.
We propose a novel, neural network-based method, which produces forecasts for all locations and lead times, jointly.
To relax the distributional assumption made by many post-processing methods, our approach incorporates normalizing flows as flexible parametric distribution estimators.
This enables us to model varying forecast distributions in a mathematically exact way.
We demonstrate the effectiveness of our method in the context of the EUPPBench benchmark, where we conduct temperature forecast post-processing for stations in a sub-region of western Europe.
We show that our novel method exhibit state-of-the-art performance on the benchmark, outclassing our previous, well-performing entry.
Additionally, by providing a detailed comparison of three variants of our novel post-processing method, we elucidate the reasons why our method outperforms per-lead-time-based approaches and approaches with distributional assumptions.

\section{Introduction}

Forecast post-processing is a crucial task when constructing skillful weather forecasts. 
This is due to the inherent biases of the numerical weather predictions (NWP) which are the results of initial condition errors, computational simplifications, and sub-grid parametrizations \cite{bauer_quiet_2015,vannitsem_statistical_2021,hakim_weather_2017}. 
All these factors combine and result in errors that compound as the forecast horizon increases. 
Therefore, due to these issues, the forecast is uncertain by nature. 
To quantify this uncertainty ensemble forecasts are issued by centers for weather forecasts, such as the European Center for Medium Range Weather Forecasts (ECMWF) \cite{ecmwf}. 
ECMWF generates ensemble forecasts by perturbing the initial atmospheric conditions and varying the NWP parametrizations. 
This results in $50$ different ensemble members, each expressing a potentially unique future weather situation. 
However, biases are still present in ensemble forecasts resulting in a disconnect between the specific observations and the described ensemble distribution.

To mitigate the effects of biases in forecasts with the aim of constructing calibrated forecast probability distributions \cite{gneiting_probabilistic_2007} forecast providers frequently employ forecast post-processing techniques \cite{vannitsem_statistical_2021}.
These statistical approaches transform raw ensemble forecasts to better concur with actual weather variable observations.  
Post-processing methods vary from simple to complex, from rigid \cite{gneiting_calibrated_2005,raftery_using_2005,schulz_post-processing_2021,phipps_evaluating_2022} to flexible \cite{kirkwood_framework_2021,hewson_low-cost_2021,moller_vine_2018,van_schaeybroeck_ensemble_2015}, and encompass a large spectra of statistical approaches \cite{vannitsem_statistical_2021}.
Another way of distinguishing between different post-processing methods is to characterize them based on their distributional assumptions or lack thereof. 
Since the nature of ensemble forecasts is inherently probabilistic (ensemble forecasts quantify the weather forecast uncertainty) the post-processing output should reflect that. 
Methods that assume the target weather variable distribution are usually simpler to implement however, they suffer from the pitfall of assuming the wrong distribution. 
This can result in additional biases. 
Likewise, different weather variables can exhibit different distributions therefore, tuning and exploration of the best target distribution is required before constructing the model. 
Methods that do not make such assumptions are usually more complex to implement and require more data for parameter estimation.

Recently neural networks started showing promising results in this field \cite{mcgovern_using_2017,lerch_convolutional_2022,chapman_probabilistic_2022,schulz_machine_2022,veldkamp_statistical_2021,bremnes_ensemble_2020,gronquist_deep_2021,scheuerer_using_2020,rasp_neural_2018,dupuy_arpege_2021,mcgovern_using_2017,gneiting_probabilistic_2023}. 
Applications such as those by \cite{rasp_neural_2018} exhibited state of the art results compared to conventional post-processing techniques. 
By using the ensemble mean and variance with the combination of station embeddings \cite{rasp_neural_2018} construct a neural network that estimates the parameters of a normal distribution for each lead time. 
This approach is further extended in \cite{schulz_machine_2022} by applying different distribution estimators in place of the normal distribution. 
To be more specific, they apply the same neural network architecture as a base parameter predictor to three different models: a distributional regression model, a Bernstein quantile regression model \cite{bremnes_ensemble_2020}, and a histogram estimation model. 
Indeed, their aim to reduce the distributional assumption is apparent and also required as many weather variables can not be effectively described using simple probability distributions. 
We can make a similar observation by investigating the work by \cite{veldkamp_statistical_2021} where the authors construct a convolutional neural network for post-processing wind speeds. 
Similarly to \cite{schulz_machine_2022}, they tried different probability models as the output of their convolutional neural network: a quantized softmax approach, kernel mixture networks \cite{ambrogioni_kernel_2017}, and truncated normal distribution fitting approach. 
\cite{veldkamp_statistical_2021} concluded that the most flexible distribution assessment method exhibited the best performance. 
This further bolsters the need for sophisticated distribution estimation techniques as these increase the performance of post-processing algorithms.

To further improve upon the existing post-processing techniques we propose two novel neural network approaches, Atmosphere NETwork 1 (ANET1) and Atmosphere NETwork 2 (ANET2).
The novelty of the ANET methods is threefold and can be summarized thusly:
\begin{itemize}
    \item Two novel neural network architectures for post-processing ensemble weather forecasts
    \item Joint probabilistic forecasting for all lead times and locations
    \item Flexible parametric distribution estimation based on normalizing spline flows, implemented in ANET2
\end{itemize}
Both ANET variants are constructed such that there is only one model for the entire post-processing region and all lead times.
This is often not the case with frequently used post-processing methods \cite{demaeyer_euppbench_2023} which typically construct individual models per-lead-time and sometimes even per-station.
Our joint location-lead-time method ANET1 exhibits state-of-the-art performance on the EUPPBench benchmark \cite{demaeyer_euppbench_2023} compared to other submitted methods, which are implemented on a per-lead time or per station basis.
\\
ANET2 is the second iteration of our post-processing approach and features an optimized neural network architecture (similar to \cite{rasp_neural_2018}) and training procedure.
Furthermore, ANET2 improves upon ANET1 by relaxing the distribution assumption of ANET1, which assumes that the target weather variable is distributed according to a parametric distribution which, in the case of temperature, is a normal distribution.
ANET2 relaxes this assumption by modeling the target distribution in terms of a normalizing spline flow \cite{durkan_neural_2019,dinh_density_2016,kobyzev_normalizing_2020,jaini_tails_2020}.
This enables us to model mathematically exact distributions without specifying a concrete target distribution family.
We demonstrate that ANET2 further improves upon ANET1 in a suite of performance metrics tailored for probabilistic forecast evaluation.
Additionally, we provide intuition as to why such a type of joint forecasting and model construction leads to better performance by analyzing the feature importance encoded in the ANET2 model.

To demonstrate the performance of the ANET variants we compare four different novel methods for probabilistic post-processing of ensemble temperature forecasts: ANET1, ANET2, ANET2$_\text{NORM}$, and ANET2$_\text{BERN}$.
This helps us quantify the impact different neural network architectures and distributional models have on the final post-processing performance of our approaches.
The description of the aforementioned methods and training procedures is available in Section \ref{sec:mthds}.
We follow this with the evaluation results in Section \ref{sec:rslts} which we further elaborate on in Section \ref{sec:dscssn}.

\section{Methods}
\label{sec:mthds}

\subsection{ANET1}
\label{sec:anet1}

\begin{figure}[ht]
    \centering
    \includegraphics[width=1.0\linewidth, valign=t]{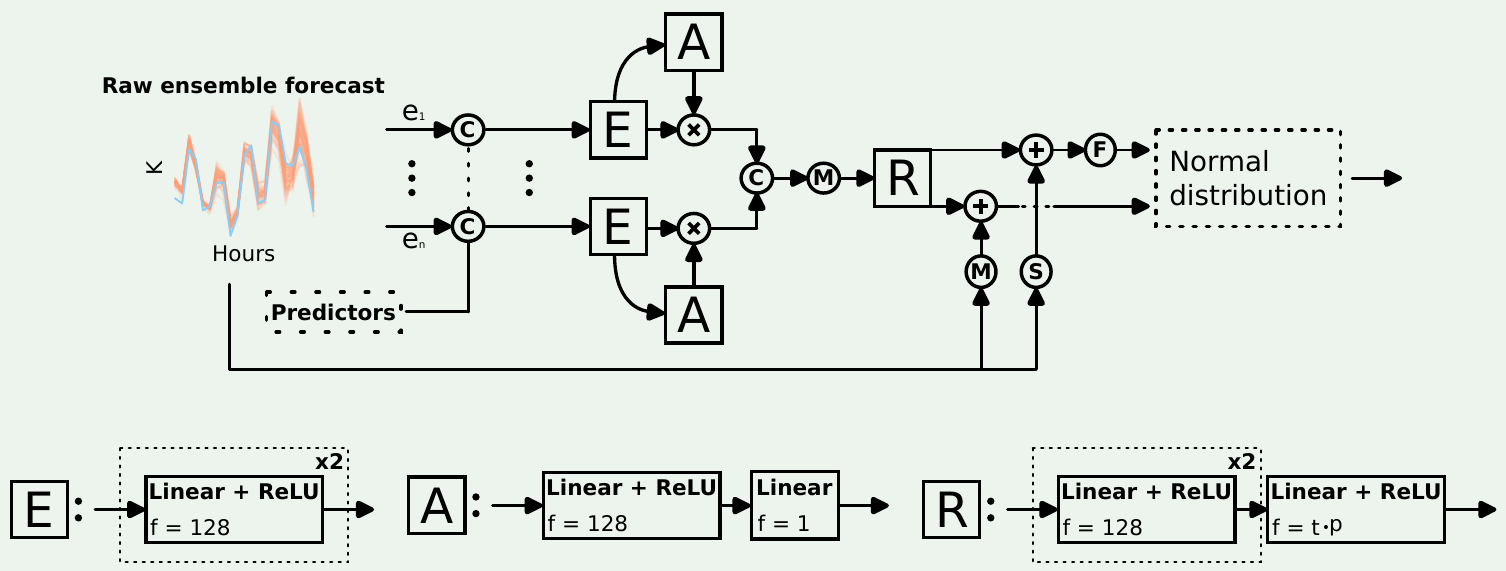}
    \caption{The neural network architecture of ANET1. The parameter f denotes the number of features in a dense layer. The parameters t and p denote the lead time and the number of per-lead-time parameters required by the normal distribution model. 
    The circular blocks denote the following operations: \textbf{M} and \textbf{S} the ensemble forecast mean and standard deviation, \textbf{X} the element-wise product, \textbf{C} concatenation operation, and \textbf{F} the softplus activation \cite{noauthor_softplus_nodate}. The square blocks denote the following operations: \textbf{E} corresponds to the shared forecast encoder block, \textbf{A} corresponds to the dynamic attention block, and \textbf{R} corresponds to the regression block. 
    The variable e$_\text{i}$ denotes the $\text{i}$-th ensemble member in the forecast.}
    \label{fig:anet1_arch}
\end{figure}

ANET1 is a neural network based approach for probabilistic ensemble forecast post-processing. 
The ANET1 architecture, displayed in Figure \ref{fig:anet1_arch}, is tailored for processing ensemble forecasts with a varying number of ensemble members. 
ANET1 achieves this by first processing ensemble forecasts for the entire lead time individually, by passing them through a shared forecast encoder structure.
We concatenate per-station predictors to each ensemble forecast to allow ANET1 to adapt to individual station conditions.
These predictors include station and model altitude, longitude, latitude, and land usage, with the addition of a seasonal time encoding, defined as $\text{cos}(\frac{2\pi d}{365}),$ where $d$ denotes the day of year the individual input forecast was issued.
The shared forecast encoder transforms each forecast-predictor pair into a high-dimensional latent encoding, which are then weighed by a dynamic attention block and averaged into a single, mean ensemble encoding.
This dynamic attention mechanism is implemented with the goal of determining the importance of individual ensemble members in a given weather situation.
The mean ensemble encoding is then passed through a regression block, the outputs of which are the per lead time additive corrections to the ensemble mean and standard deviation.
We use the corrected mean and standard deviation as parameters of the predictive normal distribution.

\subsection{ANET2}
\label{sec:anet2}

ANET1 suffers from the distributional assumption drawback which mitigates its expressive power.
To improve upon ANET1 we developed ANET2 which utilizes normalizing spline flows \cite{durkan_neural_2019} in place of the normal distribution.
Normalizing flows are designed for density estimation and can approximate complex distributions in a tractable and mathematically exact way.
ANET2 embraces this methodology, albeit in a modified manner which better suits our context of application, by conditioning the final distributional model on the provided raw ensemble forecast.

\subsubsection{Flexible parametric distribution estimation using modified normalizing flows}

We begin our overview of the normalizing spline flow density estimation procedure by defining a rational-quadratic spline transformation $\Tilde{T}(x; \theta)$, parameterized by $\theta$, with $x$ as the target temperature realization.
The transformation $\Tilde{T}_{\theta}$ is a strictly increasing and differentiable function and is the key behind the expressive power of normalizing spline flows.
The parameter set $\theta$ includes the spline knots and the spline's values at those knots.
Since our goal is to model the target temperature distribution $F_{temp}$ we can use the change of variable approach \cite{kobyzev_normalizing_2020} to express this unknown distribution in terms of the transformation $\Tilde{T}_{\theta}$ and a base distribution $F_{norm}$ which, in our case, is a univariate normal distribution.
The target variable density can then be defined as 
\begin{equation*}
p_{temp}(x;\theta) = p_{norm}(\Tilde{T}_{\theta}(x))\frac{\partial \Tilde{T}_{\theta}(x)}{\partial x}.
\end{equation*}
Since $p_{norm}$ is the density of a normal distribution with zero mean and standard deviation of one, the final loss function minimized (the negative log-likelihood) for a given sample $x_i$ is
\begin{equation*}
L(p_{temp}(x_i)) = \frac{\Tilde{T}_\theta(x_i)^2}{2} - \ln(\frac{\partial \Tilde{T}_\theta(x_i)}{\partial x}).
\end{equation*}

In practice we replace $\Tilde{T}_\theta$ with a composition of rational-quadratic spline transformations $T_\Theta$, where
\begin{equation*}
    T_\Theta = \Tilde{T}_{\theta_{l}}(\Tilde{T}_{\theta_{l - 1}}(...)),
\end{equation*}
and $\Theta$ denotes the entire set of individual spline parameters $\theta_l$.
ANET2 uses a composition of four rational-quadratic spline transformations, each described by four knot-value pairs.
Since ANET2 generates a distribution for each lead time this entails that the parametric distribution described by this model contains $840$ parameters ($21 \times 10 \times 4$: $21$ for each lead time, $10$ for each of the $4$ spline transformations).
To fully describe a rational-quadratic spline we require the spline's knot-value pairs and the value of the spline's derivative at those knots. 
The knot-value pairs are estimated by the ANET2 neural network (described in more detail in Section \ref{sec:prmtr_anet2}).
For a specific ensemble forecast with a lead time of $t$ steps, ANET2 computes $t$ parameter sets $\Theta_j$, where $j \in [1, t]$ (in our case, $t = 21$).

Each $\theta_l$ is comprised of two vectors, where the first contains the knots and the second the values of the spline.
More formally, 
\begin{equation*}
\theta_l := \{\bm{k}_l, \bm{v}_l\},
\end{equation*}
where the vector $\bm{k}_l$ denotes the spline knots and $\bm{v}_l$ the values. 
We use a neural network to estimate these values based on the input ensemble forecast. 
The monotonicity of the spline knot-value pairs is ensured using the $\text{Softplus}$ function \cite{noauthor_softplus_nodate}. 
Therefore, the final vectors of knots $\bm{k}_l$ and values $\bm{v}_l$ are defined as
\begin{align*}
\bm{k}_l &:= \text{CumSum}([\bm{k}_{l,1}', 1\text{e}^{-3} + \text{Softplus}(\bm{k}_{l, [2,5]}')]),\\
\bm{v}_l &:= \text{CumSum}([\bm{v}_{l,1}', 1\text{e}^{-3} + \text{Softplus}(\bm{v}_{l, [2,5]}')]),
\end{align*}
where $\bm{k}'_l$ and $\bm{v}'_l$ refer to the raw neural network outputs for the knot-value pairs, $\bm{k}_{l,1}'$ refers to the first element of the vector $\bm{k}_l'$, while $\bm{k}_{l, [2,5]}'$ contains the remaining elements (the same holds true for $\bm{v}_l'$. 
The index into $\bm{k}_l'$ and $\bm{v}_l'$ goes up to five as we only have five knot-value pairs per spline.
Additionally, we limit the minimal distance between two consecutive knot-value pairs to be $1\text{e}^{-3}$. 
We implement this restriction to ensure numerical stability and in our testing this does not impact the regression performance. 
The function $\text{CumSum}$ denotes the cumulative sum operation, where the first element is kept identical to that of the input vector.
This in combination with the $\text{Softplus}$ function ensures that the knots-value pairs increase monotonically, in turn guaranteeing the monotonicity of the rational quadratic spline transformation \cite{gregory_piecewise_1982}.

However, to provide a full estimate of the transformation we also require the positive derivative values at the knots, denoted as $d_{l, i}$, where $i \in [1, ..., 5]$. 
\cite{durkan_neural_2019} propose that we determine these derivatives from the neural network in much the same way as the knot-value pairs. 
This definition of the derivative can yield discontinuities in the splines and therefore in the final estimated density.
This is not practical for our application context as we frequently inspect the probability density to determine the most likely weather outcome. 
A density full of discontinuities conveys unnatural properties and impedes the required analysis. 
To rectify this we look at the work by \cite{gregory_piecewise_1982}, where they propose two derivative estimation schemes that compute the required derivatives from the knot-value pairs. 
We adapt the second approach in our work, which is expressed as
\begin{equation*}
d_{l, i} = \frac{\Delta_{l, i}\cdot\Delta_{l, i - 1}}{\delta_{l, j}}
\end{equation*}
for $i \in [2, 4]$.
The derivative values at the edge knots are defined as
\begin{equation*}
d_{l, 1} = \frac{\Delta_{l, 1}^2}{\delta_{l, 2}}, \hspace{0.2cm} d_{l, 5} = \frac{\Delta_{l, 4}^2}{\delta_{l, 4}},
\end{equation*}
where $\Delta_{l, j}$ and $\delta_{l, j}$ are computed by
\begin{equation*}
\Delta_{l, j} = \frac{\bm{v}_{l, j + 1} - \bm{v}_{l, j}}{\bm{k}_{l, j + 1} - \bm{k}_{l, j}}, \hspace{0.2cm}
\delta_{l, j} = \frac{\bm{v}_{l, j + 1} - \bm{v}_{l, j - 1}}{\bm{k}_{l, j + 1} - \bm{k}_{l, j - 1}}.
\end{equation*}
Due to our constraint on the minimal differences between consecutive knots and values we do not have to concern ourselves with specific edge cases outlined in \cite{gregory_piecewise_1982} where the differences in above equalities would be zero.

Additionally, we do not restrict the spline knots to a predetermined interval. 
This is contrary to \cite{durkan_neural_2019}, however, in our testing, this constraint relaxation works well and provides no tangible difference in performance. 
It does, however, eliminate the need for parameter tuning as the knot ranges are determined automatically during optimization.

%We estimate the parameter values of the spline transformations using gradient descent by minimizing the negative log-likelihood of the distribution model.

\subsubsection{ANET2$_{\text{NORM}}$}

ANET2$_{\text{NORM}}$ combines the ANET2 neural network model described in Section \ref{sec:prmtr_anet2} with a normal distribution acting as its probability distribution model.
Therefore, we model the mean and standard deviation vectors of the distribution, each containing $21$ values corresponding to the lead times. 
Let us denote the raw ensemble mean and standard deviation vectors as $\bm{\mu}^{\text{e}},\bm{\sigma}^{\text{e}}$. 
We compute the final model mean and standard deviation vectors $\bm{\mu},\bm{\sigma}$ as
\begin{align} 
\bm{\mu} & = \bm{\mu}^\text{e} + \bm{\mu}',\\
\bm{\sigma} & = \bm{\sigma}^\text{e} + \text{Softplus}(\bm{\sigma}'),
\end{align}
where $\bm{\mu}', \bm{\sigma}'$ denote the raw neural network estimates for those parameters. 
We again use the $\text{Softplus}$ function to enforce the positivity constraint on the standard deviation. 
We found that if we use the neural network to predict the mean and standard deviation correction residual (expressed as an additive correction term to the raw ensemble statistics) the model performs better than it would if it were to directly predict the target mean and standard deviation without the residual.
%The model is trained by minimizing the negative log-likelihood loss.

\subsubsection{ANET2$_{\text{BERN}}$}

Similarly to ANET2$_{\text{NORM}}$, we form ANET2 $_\text{BERN}$ by combining the neural network parameter estimation architecture of ANET2 with the quantile regression framework described by \cite{bremnes_ensemble_2020} as its probability distribution model.
In this case, the output of the neural network model is a set of $21$ Bernstein polynomial coefficient vectors, each containing $13$ values. 
Therefore, the degree of the Bernstein polynomial we fit is $12$ (the degree is one less than the number of parameters). 
As per the suggestion of \cite{bremnes_ensemble_2020} we train the model by minimizing the quantile loss on $100$ equidistant quantiles. 
To ensure that no quantile crossing can occur, \cite{bremnes_ensemble_2020} suggest one might restrict the polynomial coefficient to the positive reals.
However, in our testing, this limits the expressive power of the model. 
Therefore, we let the coefficients be unconstrained as the quantile crossing event is rare in practice \cite{bremnes_ensemble_2020} and we did not encounter it in our testing (however, one must not neglect this correctness concession as it can be a source of potential issues if left unchecked).

\subsubsection{Parameter estimation neural network architecture}
\label{sec:prmtr_anet2}

\begin{figure}[ht]
    \centering
    \includegraphics[width=1.0\linewidth, valign=t]{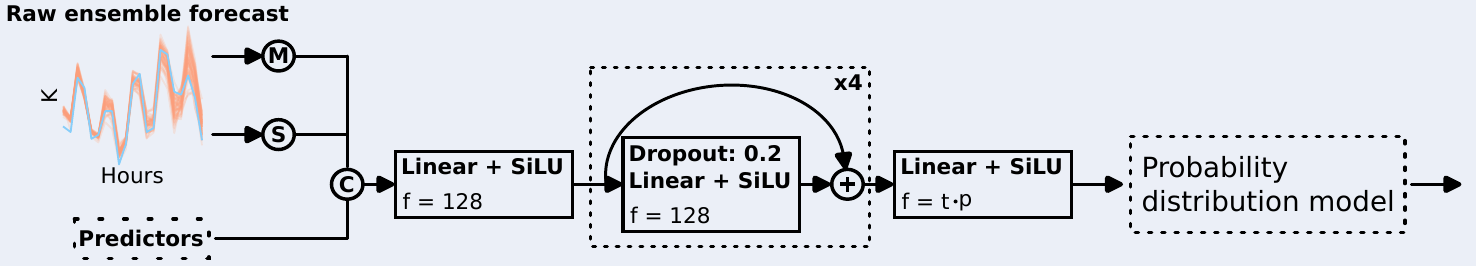}
    \caption{The neural network architecture of ANET2. The parameter f denotes the number of features in a dense layer. The parameters t and p denote the lead time and the number of per-lead-time parameters required by the normalizing flow. The circular blocks denote the following operations: \textbf{M} denotes the computation of the ensemble forecast mean, \textbf{S} block denotes the computation of the standard deviation of the ensemble forecast, and \textbf{C} denotes a concatenation operation.}
    \label{fig:anet2_arch}
\end{figure}

To determine the values of these spline parameters (in the case of ANET2$_{NORM}$ the parameters of a normal distribution and in the case of ANET2$_{BERN}$ the coefficients of the Bernstein polynomial) we use a dense neural network, whose architecture is displayed in Figure \ref{fig:anet2_arch}. 
ANET2 conducts post-processing jointly for the whole lead time and all spatial locations.
Therefore, the input to the network is an ensemble forecast with $m$ ensemble members, each containing $21$ forecasts (we will discuss the training dataset and training procedure in the following section).
ANET2 first computes the ensemble forecast mean and variance. 
These two vectors, each containing $21$ elements, are then concatenated with the per-station predictors and seasonal time encodings to form the input to the neural network.
The per-station predictors include station and model altitude, longitude, latitude, and land usage.
The seasonal time encoding is defined as $\text{cos}(\frac{2\pi d}{365}),$ where $d$ denotes the day of year the forecast was issued.
There are a total of six dense layers in the ANET2 neural network.
All dense layers, except for the last one, are followed by a SiLU \cite{hendrycks_gaussian_2020} activation, and each of the layers, except for the first and last, are preceded by a dropout layer \cite{srivastava_dropout_2014} with a dropout probability of $0.2$. 
Each of these dense layers is a residual layer implying that the transformation done by the dense layer and the consequent activation is added to the input of that layer, which then forms the output of that residual block (inspect Figure \ref{fig:anet2_arch} for more details). 
The final layer produces $21$ sets of parameters $\theta$, representing the parametric distribution for each forecast time.

\subsection{Training procedure}
\label{sec:training}

The training procedure is different between ANET1 and the remaining ANET2 variants. 
The reason for this discrepancy was the fact that ANET1 preceded ANET2 in development. 
We describe both approaches and give our arguments as to why we modified the
training procedure with ANET2.

In both cases, we used the datasets provided by the EUPPBench benchmark \cite{demaeyer_euppbench_2023} to construct the model. 
The pilot study focused on the post-processing of temperature forecasts for stations in a limited region in Europe. 
This study included two datasets.
The first dataset, denoted as $D_{11}$, consisted of $20$ year \textit{re-forecasts} \cite{demaeyer_euppbench_2023} for the years $2017, 2018$, with $209$ time samples for both years.
The ensemble for this dataset counted $m = 11$ members. 
The $D_{11}$ dataset was the designated training dataset. 
The second dataset, denoted as $D_{51}$, included \textit{forecasts} for $2017, 2018$, with $730$ time samples for both years. 
The ensemble for this dataset counted $m = 51$ members. 
The $D_{51}$ dataset was the designated test dataset. 
We performed post-processing for $229$ stations which were equal for both datasets. 
Each training batch consisted of $256$ randomly selected samples across all stations, years, and time samples. 

The datasets described above and their designated uses were equal across ANET1 and all ANET2 variants. 
The main difference between the training procedures of ANET1 and the ANET2 variants lies in how we split the $D_{11}$ dataset into the training and validation subsets.

\subsubsection{ANET1 training procedure}

To create the training-validation split we randomly partitioned the $ D_{11}$ dataset across all stations, years, and time samples. 
The first partition was the final training dataset subset, consisting of $80$ percent of $ D_{11}$. 
The remaining $20$ percent formed the validation dataset.
We select the model that exhibits the lowest validation loss (the loss function being the negative log-likelihood) as the final candidate model.
Additionally, the model did not use the validation or test datasets for parameter estimation during training.
This way we select the model that minimizes the unseen data loss in a particular training setup.
The random partitioning required us to implement early stopping conditions as otherwise the training quickly lead to overfitting due to the training-validation dataset distribution similarity.
Gradient descent was our optimization procedure of choice in conjunction with the Adam optimizer \cite{kingma_adam_2017} in the Pytorch \cite{noauthor_pytorch_nodate} framework, with its default parameters, a batch size of $ 128$ samples, learning rate of $ 10^{-3}$, and weight decay of $10^{-9}$. 
If the validation loss did not improve for more than $20$ epochs the
training was terminated.
We reduced the learning rate by a factor of $0.9$ after the validation loss plateaued for $10$ epochs to increase numerical stability.

Even though ANET1 managed to perform well in the final evaluation against competing methods \cite{demaeyer_euppbench_2023}, this training set-up poorly reflected the nature of the train-test relationship.
It also lead to stability issues.
We rectified this in the next iteration of the training procedure for ANET2.

\subsubsection{ANET2 variants training procedure}

To rectify the issues of the previous training approach we forwent the random split approach. 
Therefore, we formed the validation subset of $D_{11}$ with the re-forecasts corresponding to the year $2016$ while we used all the remaining data to train the model. 
This better reflects the train-test dynamic of $D_{11}$ and $D_{51}$, measuring the "out-of-sample" predictive power of the model, as the validation subset is not sampled from the same years as the training one.
Similarly to ANET1, we select the model that exhibits the lowest validation loss (ANET2 and ANET2$_{\text{NORM}}$ minimize the negative log-likelihood, while ANET2$_{\text{BERN}}$ is trained minimizing the quantile loss) as the final candidate model, with not using the validation and test datasets for parameter estimation during training.
This way we select the model that minimizes the unseen data loss in a particular training setup however, the training procedure is now more stable and less prone to overfitting.
The loss functions were minimized using gradient descent and the Adam optimizer in the Pytorch framework with its default parameters, a batch size of $256$ samples, learning rate of $10^{-3}$, and weight decay of $10^{-6}$.
A bigger weight decay helped with the stability of the learning procedure as ANET2's distribution modeling approach is more flexible compared to ANET1's and therefore required additional regularization.
Finally, we also reduced the learning rate by a factor of $0.9$ after the validation loss plateaued for $10$ epochs to increase numerical stability.

\section{Results}
\label{sec:rslts}

In this Section we present the evaluation of ANET1, ANET2, ANET2$_{\text{NORM}}$, ANET2$_{\text{BERN}}$ on the $D_{51}$ test dataset.
We evaluate the performance of the above methods using the continuous ranked probability score (CRPS) \cite{gneiting_probabilistic_2007}, bias, quantile loss (QL), and quantile skill score (QSS) \cite{bremnes_ensemble_2020}.

We forwent the direct comparison of ANET variants against conventional post-processing approaches.
ANET1 was already evaluated in detail against other methods in \cite{demaeyer_euppbench_2023}.
The conclusion from that comparison was that, while most post-processing methods performed similarly, ANET1 achieved the lowest CRPS and further attenuated the error variability in the day-night cycle.
ANET1's advantage compared to other approaches in terms of CRPS is most prominent at high-altitude stations (stations with altitudes bigger than $1000$ meters).
For more details, about this comparison and included method, please refer to \cite{demaeyer_euppbench_2023}.

Our evaluation of the ANET variants is displayed in Figure \ref{fig:rank_hist} and Table \ref{tbl:perf}.
In the case of QSS, we use ANET1 as the reference model, quantifying the performance gain of ANET2 variants relative to ANET1.
We compute the CRPS and bias by averaging them across all stations and time samples, resulting in a per-lead-time performance report.
The performance values in Table \ref{tbl:perf} are a result of averaging the corresponding metrics across the entire test dataset.

\subsection{Impact of improved network architecture and training procedure}

To quantify the contributions of the ANET2 neural network architecture and training procedure, we look at the performance differences between ANET1 and the ANET2 variants.
All ANET2 variants perform better on average than ANET1 across all lead times in terms of CRPS and QSS.
The difference between individual ANET2 variants is smaller, with ANET2 leading the pack.
We notice that ANET2$_{\text{NORM}}$ constantly outperforms ANET1 both in terms of average CRPS and QSS, which is further corroborated by the results in Table \ref{tbl:perf}.
However, both predict the same target distribution, that being the normal distribution.
These performance enhancements of ANET2$_\text{NORM}$ can be attributed to the architecture and training protocol changes from ANET1.
ANET2$_{\text{NORM}}$ also exhibits the lowest bias amongst all methods.
This could be due to the symmetric nature of the target normal distribution ANET2$_{\text{NORM}}$.
Conversely, ANET1 outputs the same distribution as ANET2$_{\text{NORM}}$ but has a higher bias.
We believe that this difference in bias is due to the less stable and less aligned training procedure of ANET1, which results in a sub-optimal convergence.

\subsection{Forecast calibration}

To further quantify the probabilistic calibration of individual methods we use rank histograms \cite{hamill_interpretation_2001}, displayed in Figure \ref{fig:rank_hist}.
The results are aggregated across all stations, time samples, and lead times, for each quantile. 
This gives us an estimate of how well the individual methods describe the distribution of the entire $D_{51}$ dataset.
ANET1 and ANET2$_\text{NORM}$ exhibit similar rank histograms due to both using a normal distribution as their target, implying similar deficiencies.
The standout models are ANET2, which uses the modified normalizing flow approach, and ANET2$_\text{BERN}$, which is based on Bernstein polynomial quantile regression.
Both methods are much more uniform compared to the remaining alternatives.
For example, ANET2$_\text{NORM}$ and ANET1 seem to suffer from over-dispersion as is implied by the hump between the $20$-th and $40$-th quantiles.
We can observe similar phenomena with other post-processing methods that fit a normal distribution to temperature forecasts \cite{demaeyer_euppbench_2023}.
ANET2 and ANET2$_\text{BERN}$ do not exhibit this specific over-dispersion, albeit with a small over-dispersion between the $10$-th and $30$-th quantile.
Additionally, both methods under-predict upper quantiles, albeit to a lesser degree than ANET1 and ANET2$_\text{NORM}$.
We also found that the ANET2$_\text{BERN}$ method exhibits artefacts that are most pronounced at the low and high quantiles.
We believe that this is due to the nature of the Bernstein quantile regression method where the fixed degree of the polynomial affects its expressive value, especially on the quantile edges.
Overall, ANET2 displays the most uniform rank histogram out of all evaluated methods.

\subsection{Per-altitude and per-station performance}

\begin{figure}[ht]
    \centering
    \includegraphics[width=0.32\linewidth]{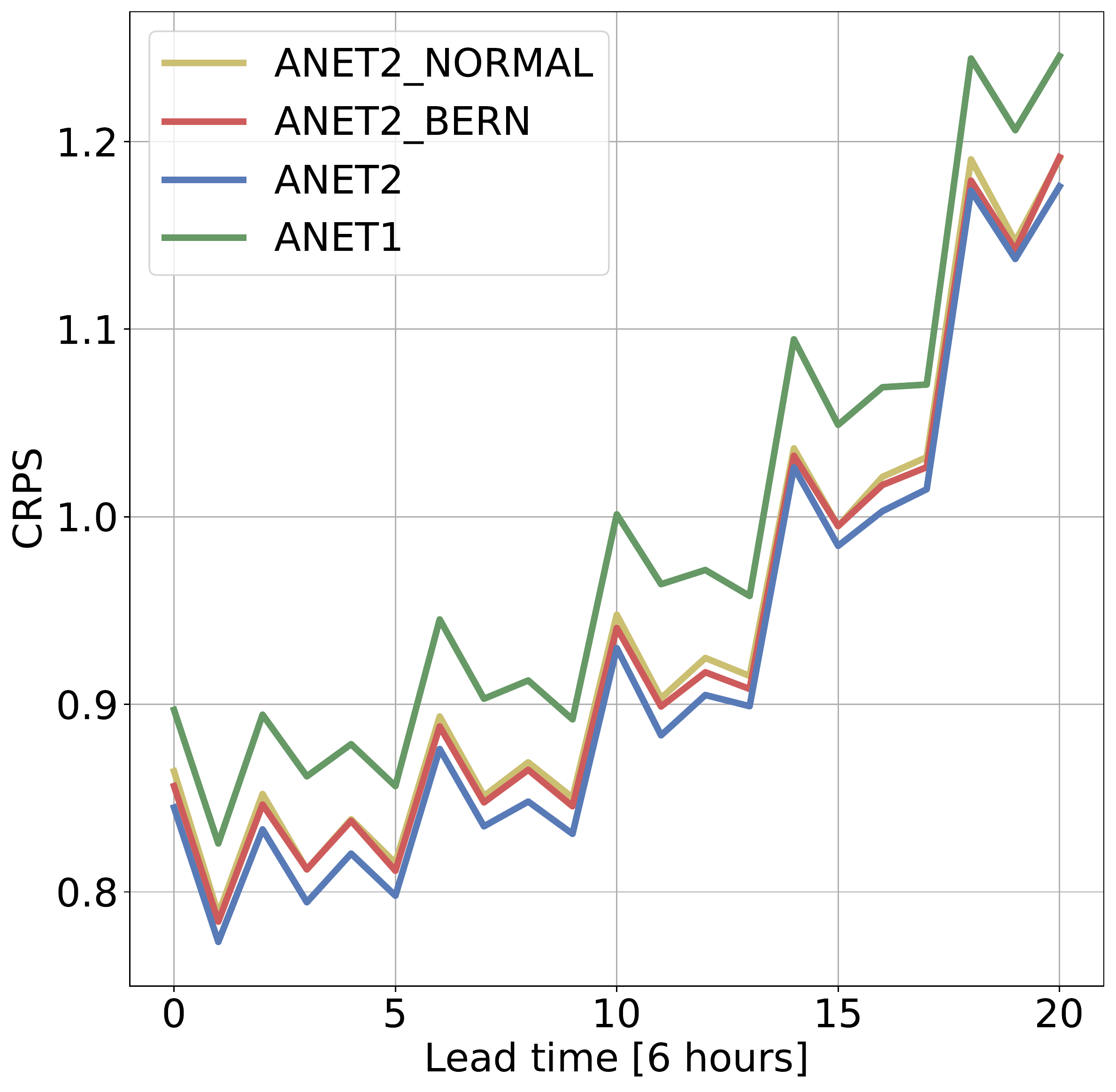}
    \includegraphics[width=0.32\linewidth]{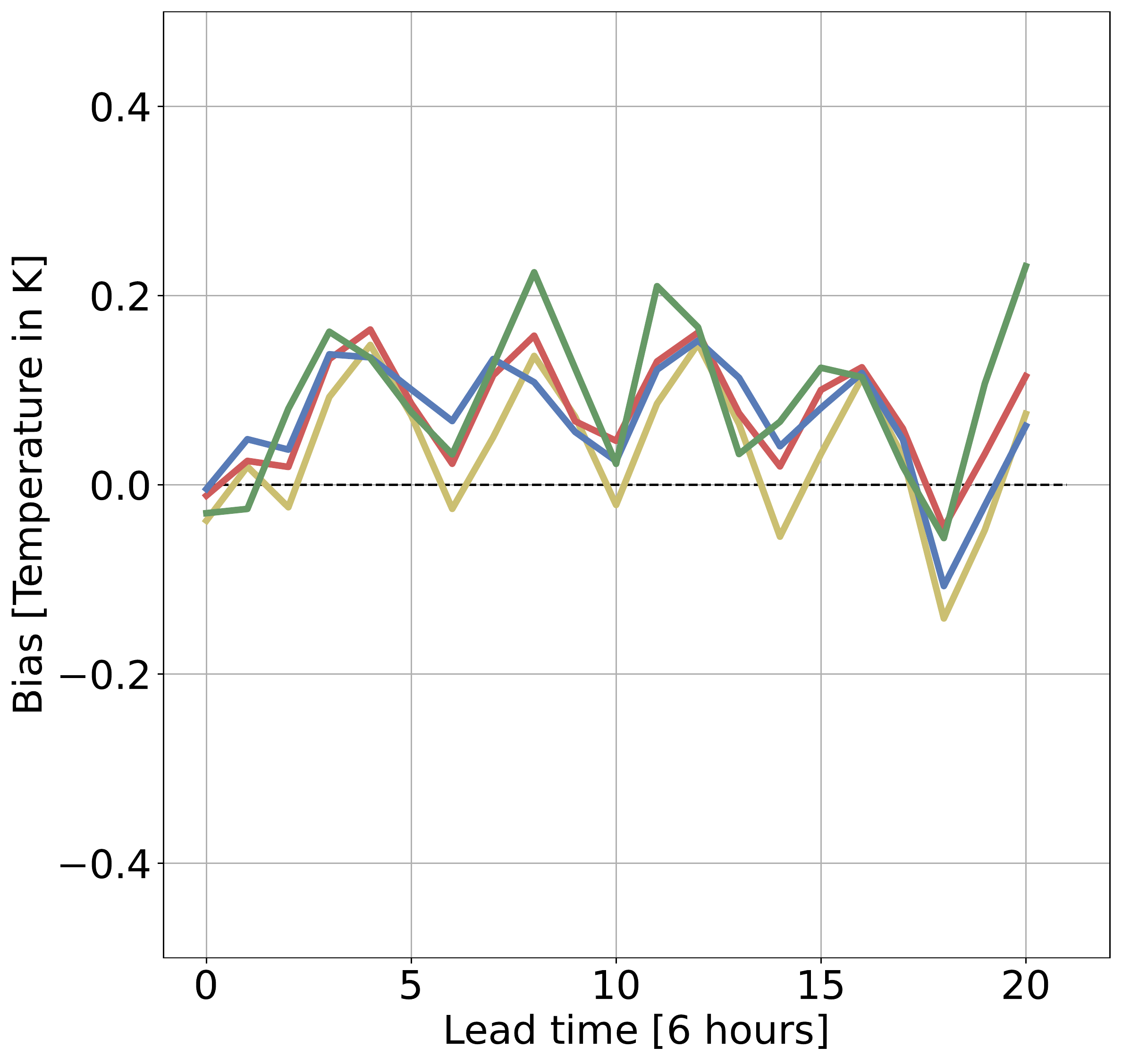}
    \includegraphics[width=0.32\linewidth]{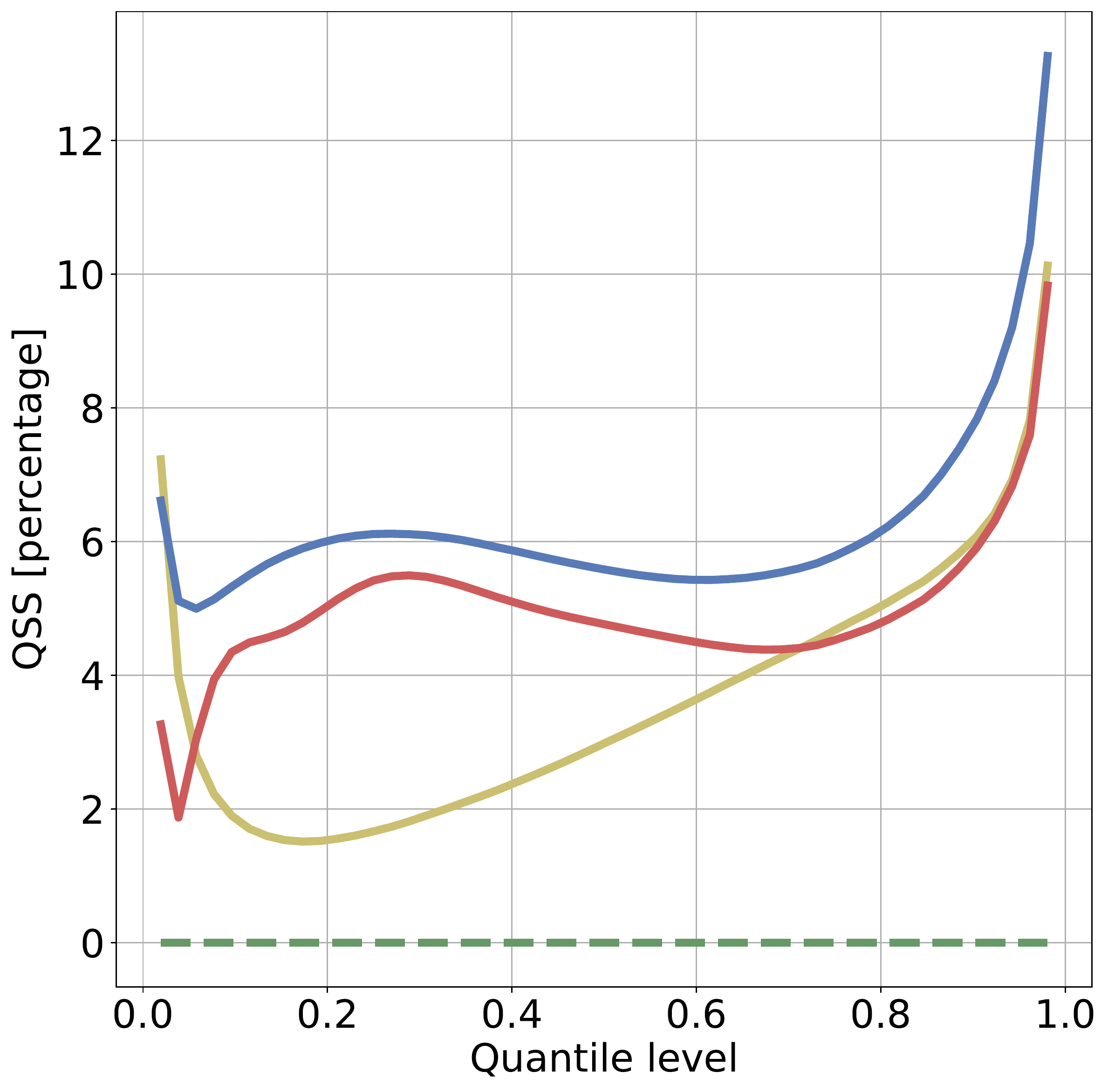}
    \includegraphics[width=0.24\linewidth]{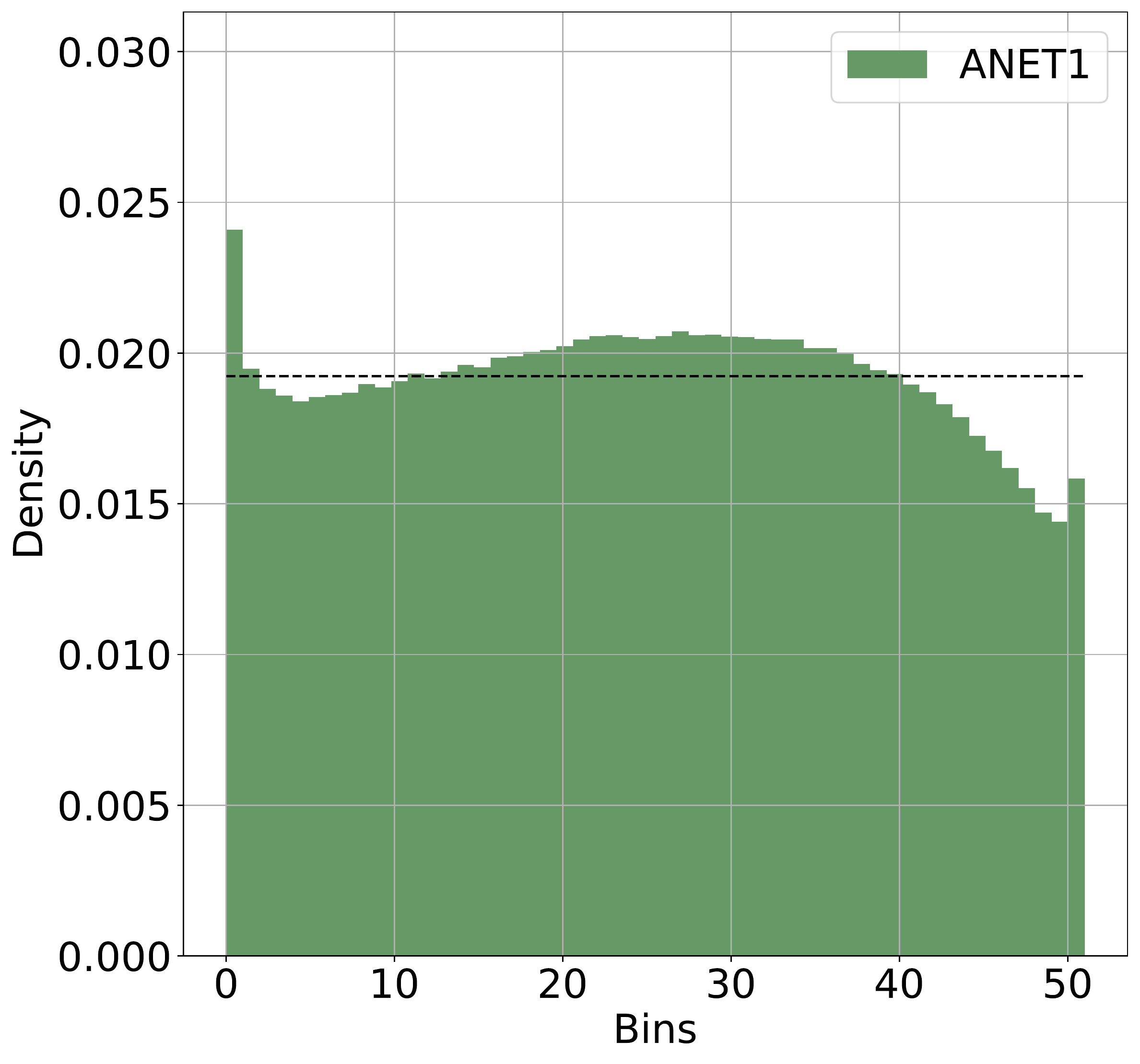}
    \includegraphics[width=0.24\linewidth]{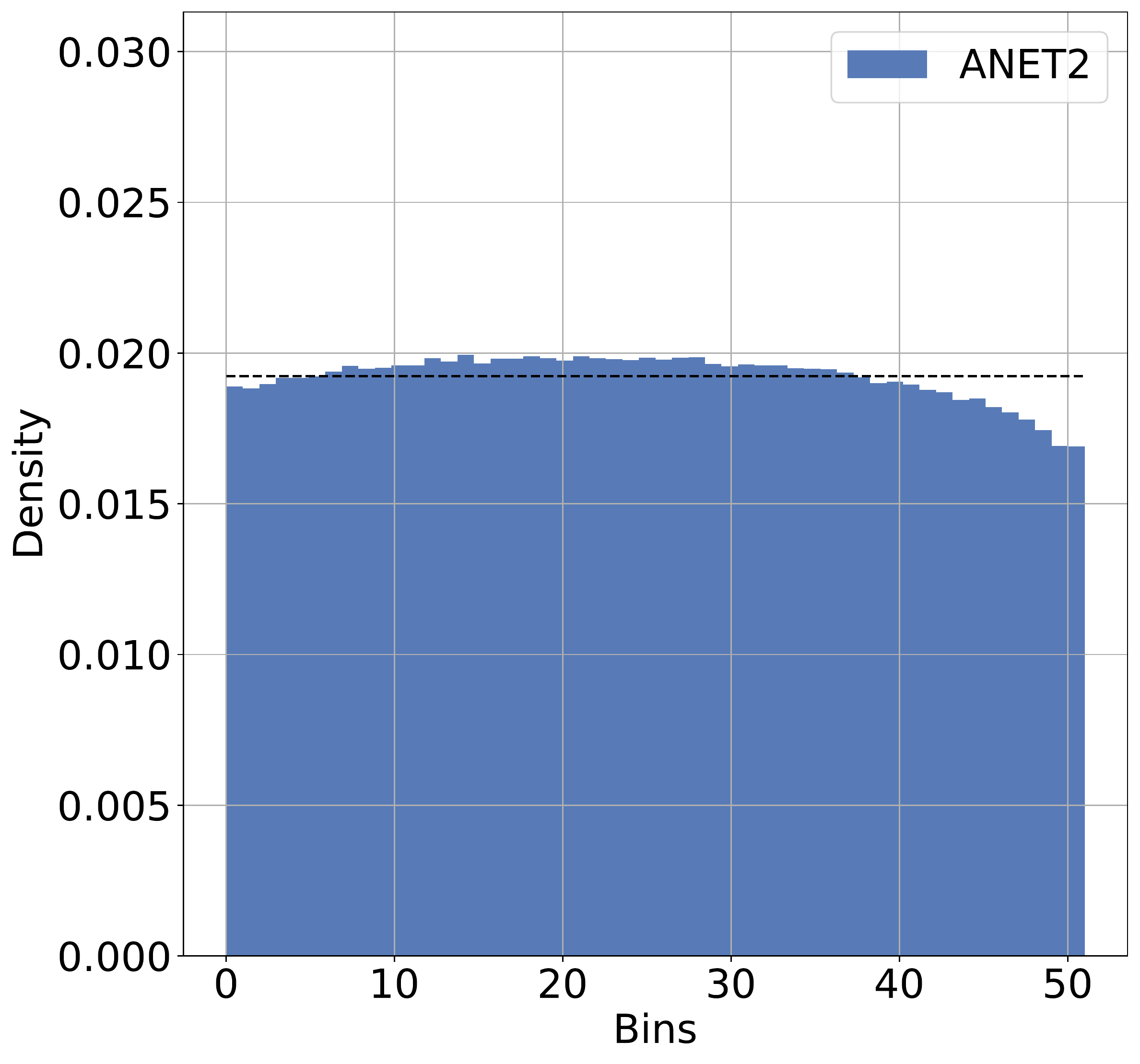}
    \includegraphics[width=0.24\linewidth]{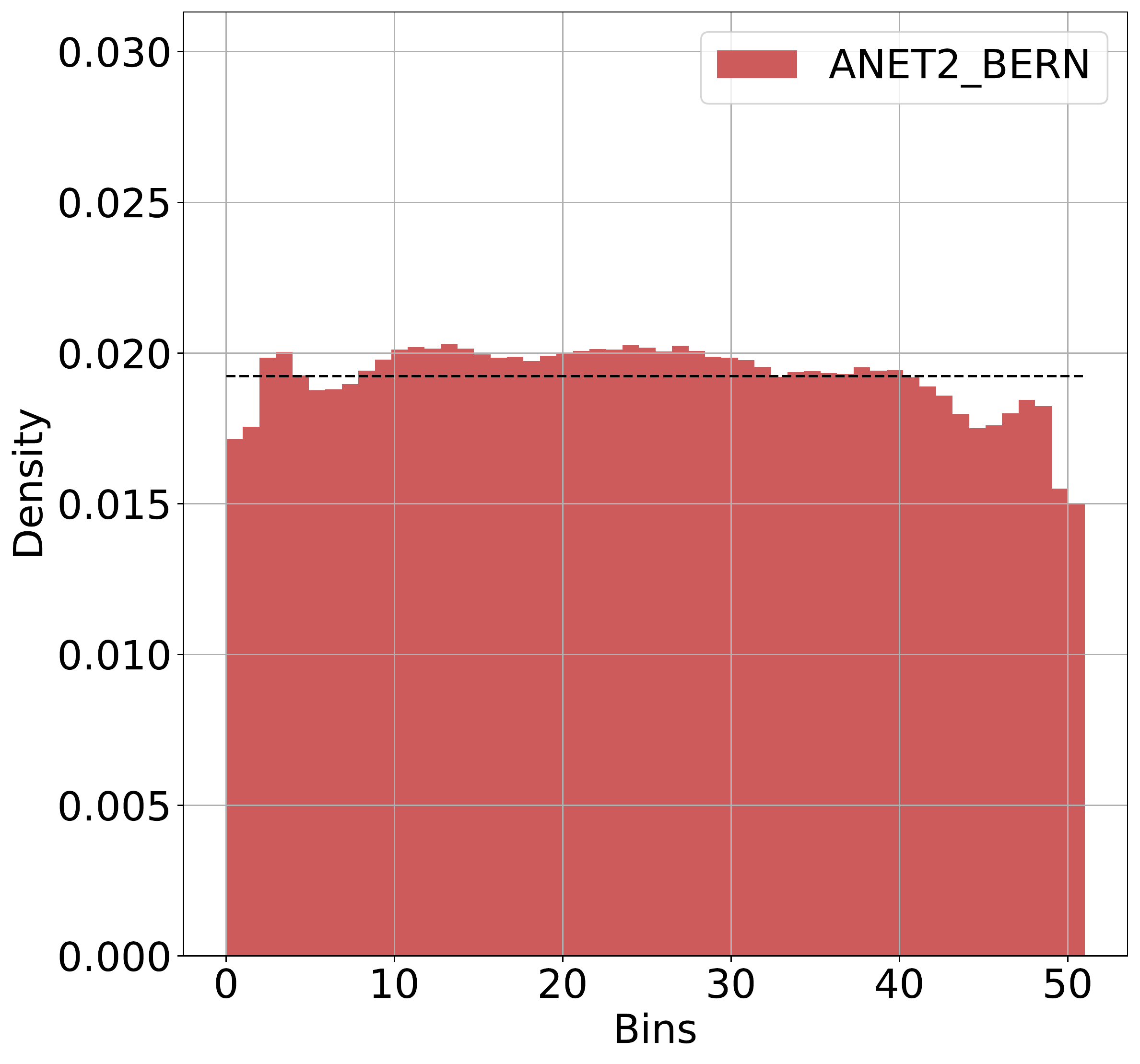}
    \includegraphics[width=0.24\linewidth]{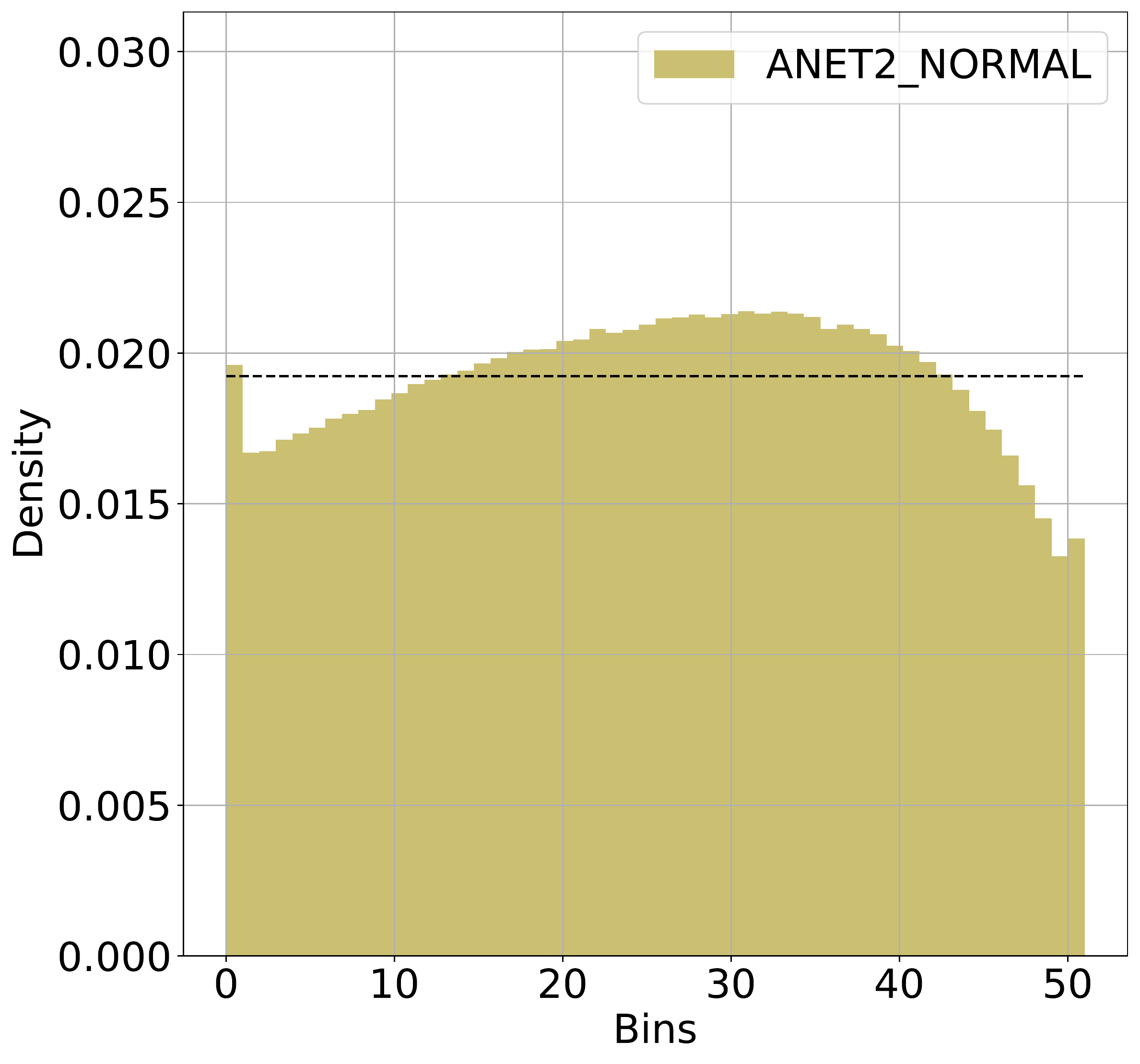}
    \caption{(Top row): The CRPS, bias, and QSS for all ANET variants. A lower CRPS is better, a bias value closer to zero is better, and a higher QSS is preferable.
    (Bottom row): The rank histogram of all ANET variants. A more uniform histogram (column heights closer to the black dashed line) is better. Histograms with central humps imply over-dispersion while histograms with outliers on the edges represent under-dispersion.}
    \label{fig:rank_hist}
\end{figure}

\begin{figure}[ht]
    \centering
    \includegraphics[width=0.32\linewidth]{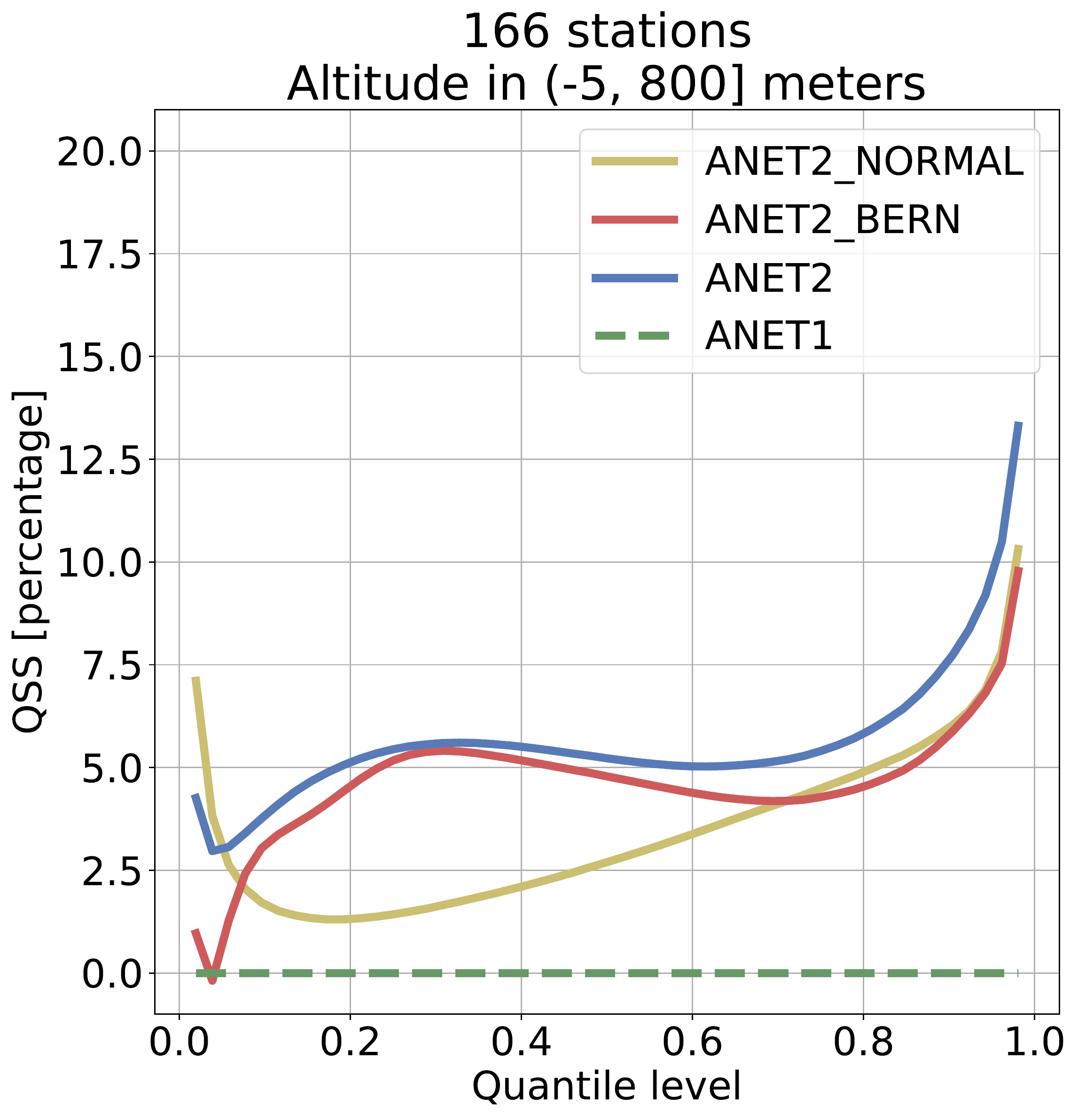}
    \includegraphics[width=0.32\linewidth]{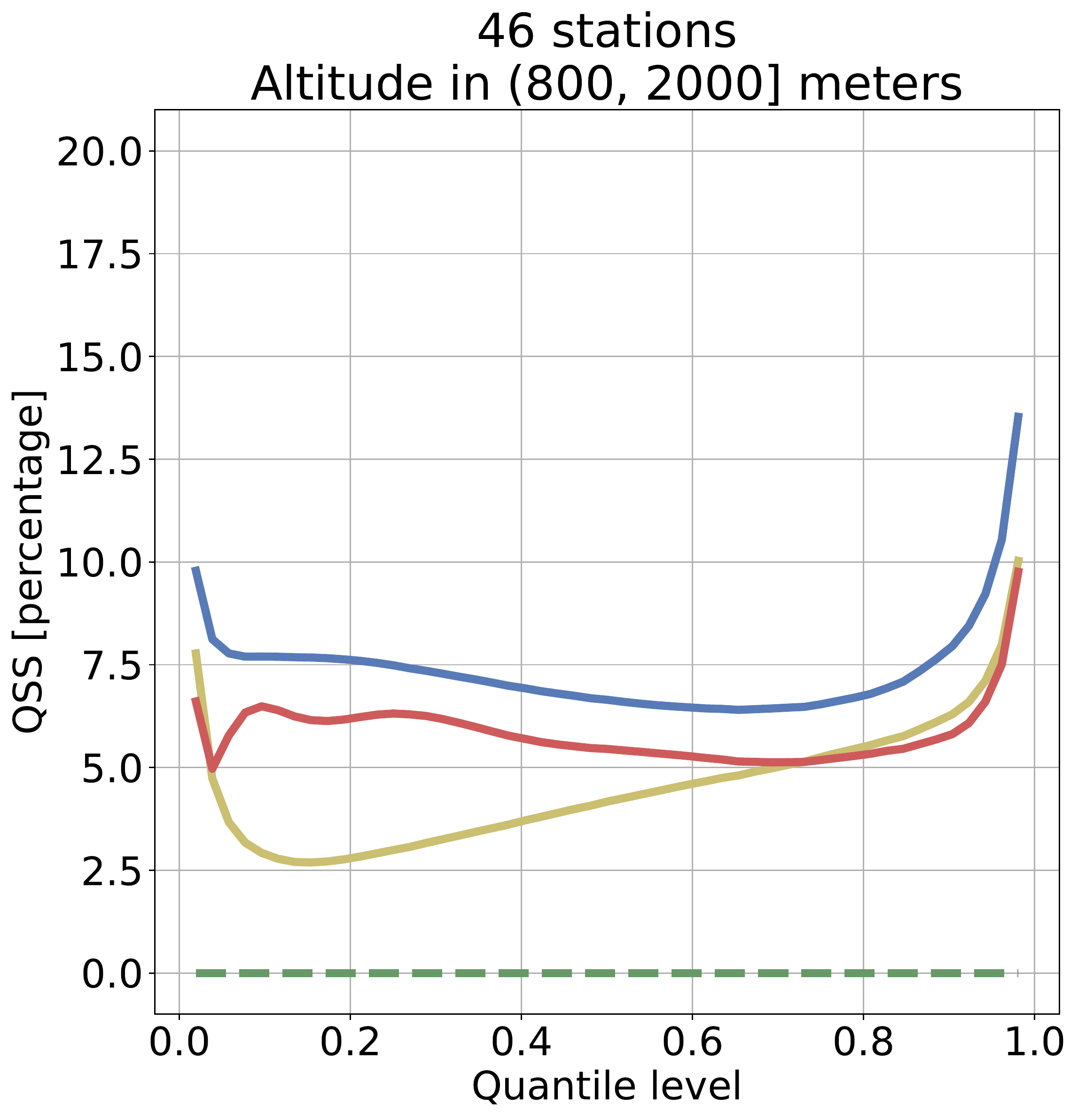}
    \includegraphics[width=0.32\linewidth]{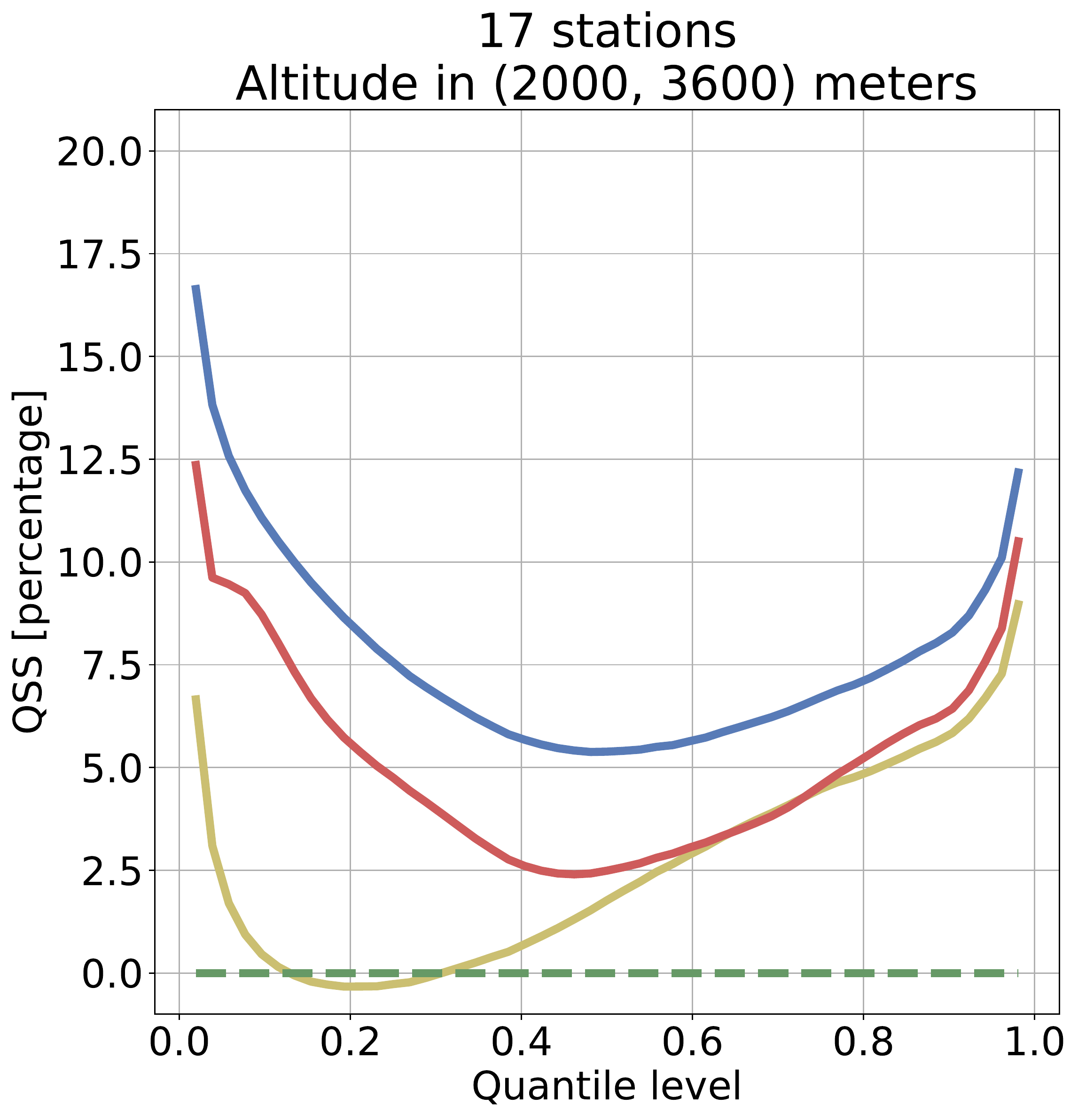}
    \caption{QSS for all compared methods relative to stations with different altitudes. ANET1 is used as the reference model, therefore, the $y$-axis denotes the percentage improvement compared to ANET1.}
    \label{fig:qss_alt}
\end{figure}

\begin{figure}[ht]
    \centering
    \includegraphics[width=0.5\linewidth, valign=t]{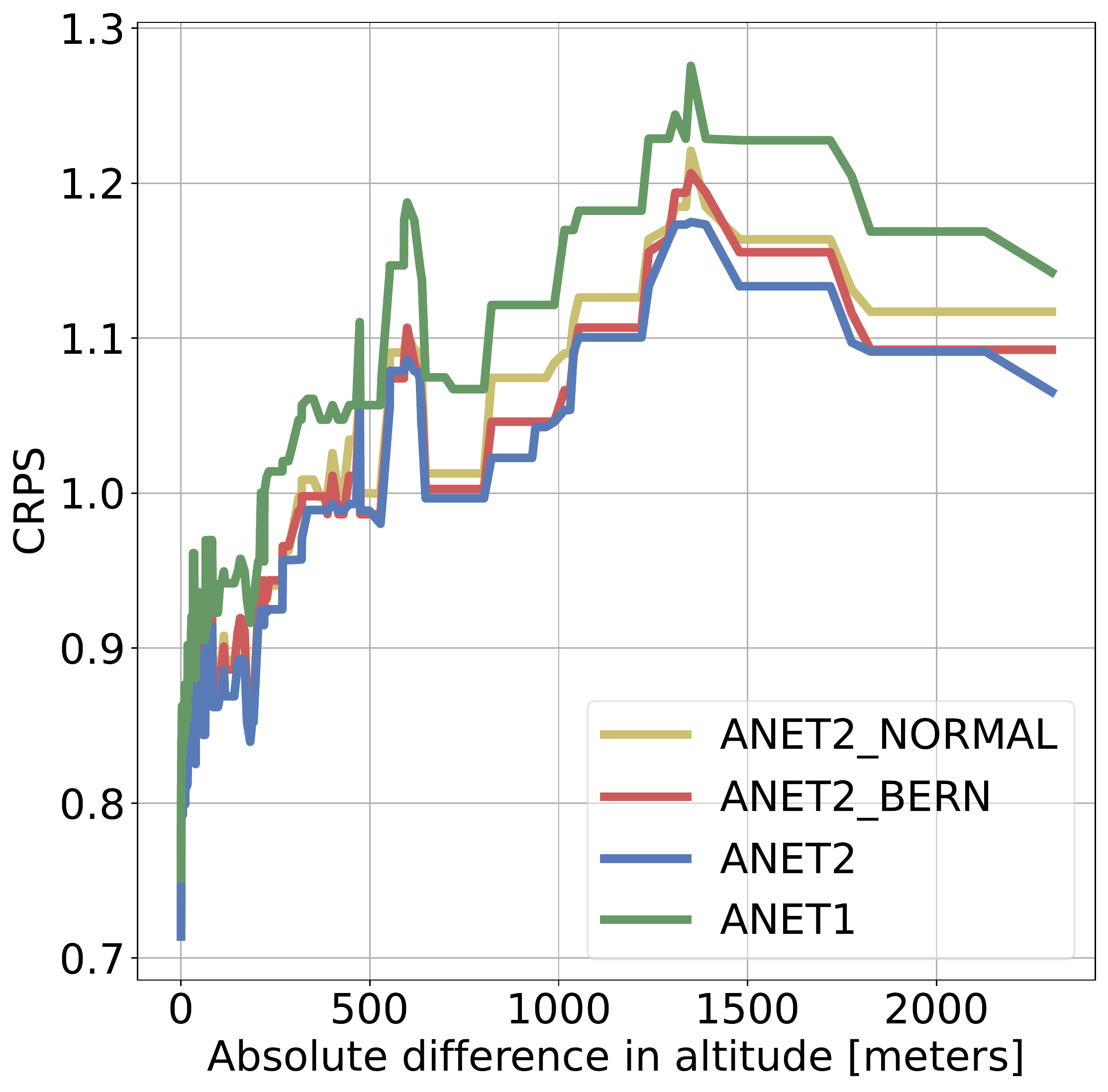}
    \includegraphics[width=0.4585\linewidth, valign=t]{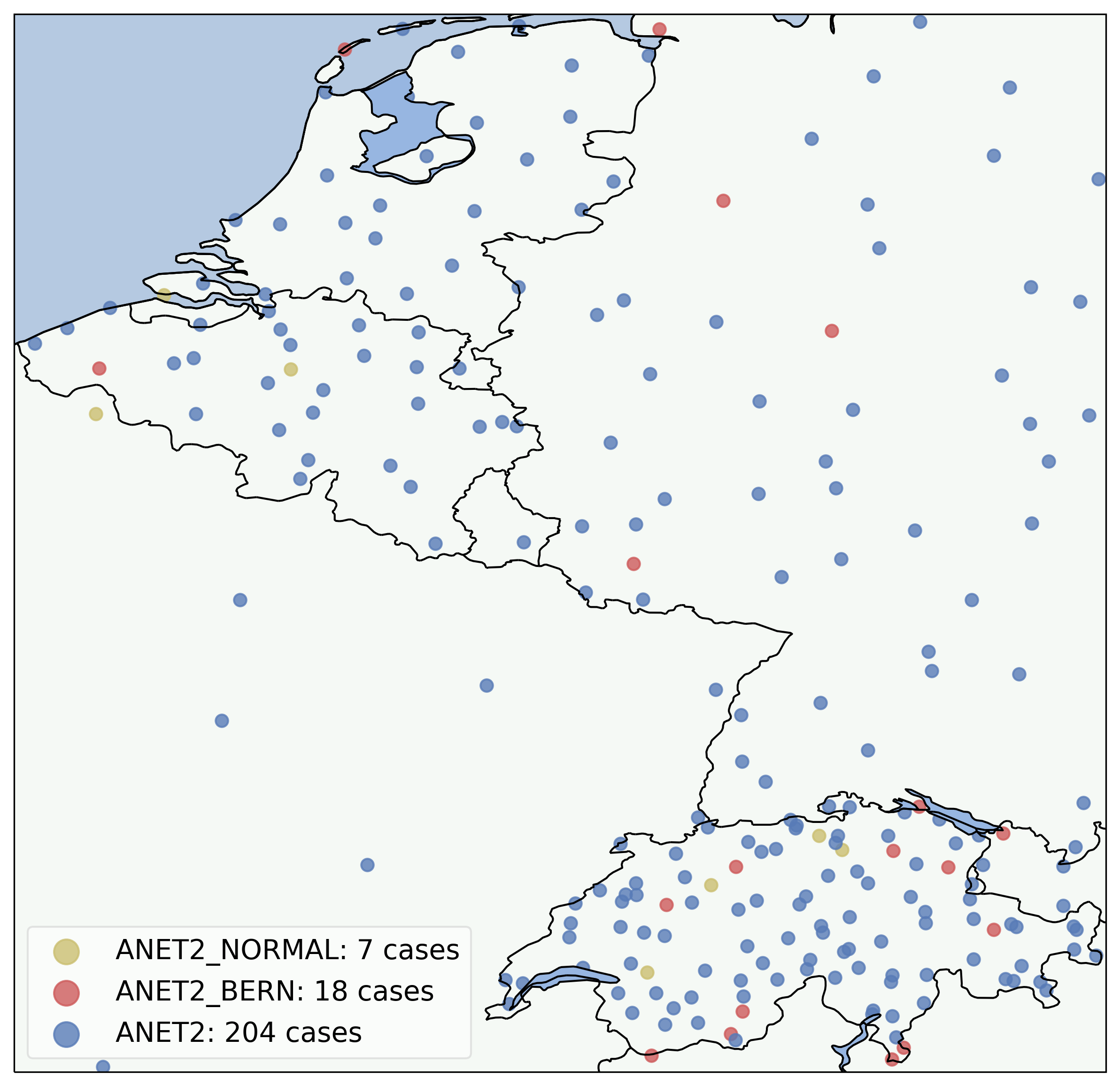}
    \caption{(Left): Average per-station CRPS relative to the absolute difference between the model and station altitude. We used median filtering with a kernel size of $15$ to suppress outliers. An increase in CRPS correlates with an increase in the absolute difference in altitude.
    (Right): Average per-station CRPS. The number of cases in which a specific method outperformed the remaining is denoted in the legend. ANET1 is not displayed since it never performed better versus the ANET2 variants.}
    \label{fig:crps_station}
\end{figure}

Station altitude correlates with the decrease in model predictive performance, as we show in Figure \ref{fig:crps_station}.
Therefore, we investigate the altitude-related performance of each model in terms of QSS over three station altitude intervals: $(-5, 800]$, $(800, 2000]$, and $(2000, 3600)$, displayed in Figure \ref{fig:qss_alt}.
We compute the per-altitude QSS by aggregating stations whose altitudes fall into a specific interval.
Then we average the QL across all lead times, time samples, and corresponding stations, producing the QL on a per-quantile level.
We use ANET1 as the reference model for the QSS.
Looking at Figure \ref{fig:qss_alt}, we can see that all ANET2 variants improve upon ANET1.
The only exceptions are ANET2$_{BERN}$ and ANET2$_{NORM}$ that exhibit roughly equal QL to ANET1 for certain quantile levels at the altitude intervals of $(-5, 800]$ and $(2000, 3600)$ respectively.
ANET2 exhibiting the highest QSS of all methods.
Indeed, ANET2 outperforms all the remaining methods over all quantile levels and all altitudes, with the exception of lower quantile levels at the stations with the lowest altitude where ANET2$_\text{NORM}$ performs the best.
The closest method, exhibiting similar trends in improvement to ANET2, is ANET2$_\text{BERN}$.
In our opinion this and the results we show in Figure \ref{fig:rank_hist} further bolster the value of flexible distribution estimators.
Similarly, by observing the average CRPS ranking per station, shown in Figure \ref{fig:crps_station}, we can see that ANET2 exhibits the best CRPS across $204$ stations from a total of $229$ stations.
ANET2$_\text{BERN}$ is the most performant method at $18$ stations, with ANET2$_\text{NORM}$ leading in CRPS only at $7$ locations.
ANET1 is not displayed as its CRPS does not outperform any of the ANET2 variants at any of the stations.
\begin{table}[ht]
\begin{center}
\begin{tabularx}{0.835\textwidth}{X | X | X | X }
  & CRPS & Bias & QL \\
\hline
ANET2$_{\text{NORM}}$ & 0.940 & \textbf{0.038} & 0.373 \\
ANET2$_{\text{BERN}}$ & 0.935 & 0.076 & 0.367 \\
ANET2                 & \textbf{0.923} & 0.069 & \textbf{0.363} \\
ANET1                 & 0.988 & 0.092 & 0.386 \\
\end{tabularx}
\end{center}
\caption{CRPS, bias, and QL (quantile loss) for all evaluated methods averaged over all stations, time samples, and lead times for the $D_{51}$ test dataset. The values in bold represent the best results for each metric.}
\label{tbl:perf}
\end{table}

\section{Discussion}
\label{sec:dscssn}

The results we presented provide a clear argument for the use of neural networks in conjunction with flexible distribution estimation methods.
Even in the case of the initial run of the EUPPBench dataset with a limited number of predictors and one target variable, temperature, neural networks can tangibly increase the state-of-the-art performance in post-processing.
The comparative study we conducted between ANET1, the improved ANET2, and its variants, revealed useful guidelines for future method research and development.

\subsection{ANET1 versus ANET2}

First, our results corroborate the findings and practices outlined by \cite{rasp_neural_2018}.
ANET1 uses a per-ensemble-member dynamic attention mechanism with the idea of determining the individual ensemble member importance, conditioned on the predictors and weather situation.
While the idea is sound in our opinion, further testing with ANET2 revealed that performing regression on the ensemble mean and variance resulted in equal or even slightly better performance, with a similar number of model parameters.
This result is similar to many existing post-processing approaches \cite{rasp_neural_2018}.
We also evaluated the effects of including additional forecast statistics into the input, such as the minimal, maximal, and median temperature.
We found that these do not impact the model's predictive performance.
We conclude that for the EUPPBench dataset v1.0 the ensemble statistics such as the mean and variance contain enough information for the formation of skillful probabilistic forecasts of temperature.
However, it likely that we have not yet found an efficient way of extracting information from individual ensemble members.

\subsection{ANET2 versus ANET2$_\text{NORM}$ and ANET2$_\text{BERN}$}

ANET2$_{\text{NORM}}$ operates by assuming the target weather distribution.
However, this method is outperformed by the remaining two, more flexible methods when observing the mean QSS and CRPS.
Similar observations were found by other researchers \cite{schulz_post-processing_2021,veldkamp_statistical_2021} further underlining the need for flexible distribution estimators.
From the two flexible approaches we implemented, ANET2 outperforms ANET2$_\text{BERN}$ in all metrics.
The ability to model the probability density through a cascade of spline transformations seems to offer greater flexibility compared to the Bernstein polynomial approach.
Additionally, ANET2 does not suffer from the quantile crossing issue that can occur with Bernstein quantile regression \cite{bremnes_ensemble_2020}.
ANET2 produces a probabilistic model from which we can estimate the exact density, distribution, and samples from the distribution.

\subsection{Joint lead time forecasting}

Where we believe that ANET2 and ANET1 innovate compared to previously applied neural network approaches is in the post-processed forecast lead time.
All ANET variants perform predictions for the entire lead time.
They are single models for all lead times and all stations, taking as input the entire ensemble forecast (whole lead time).
This enables ANET to model dependencies between different forecast times.
These ideas are further corroborated by investigating the feature importance generated using input feature permutation \cite{fisher_all_2019}.
We investigate how different lead times in the input help form the post-processed output for other lead times.

\begin{figure}[!ht]
    \centering
    \includegraphics[width=0.49\linewidth, valign=t]{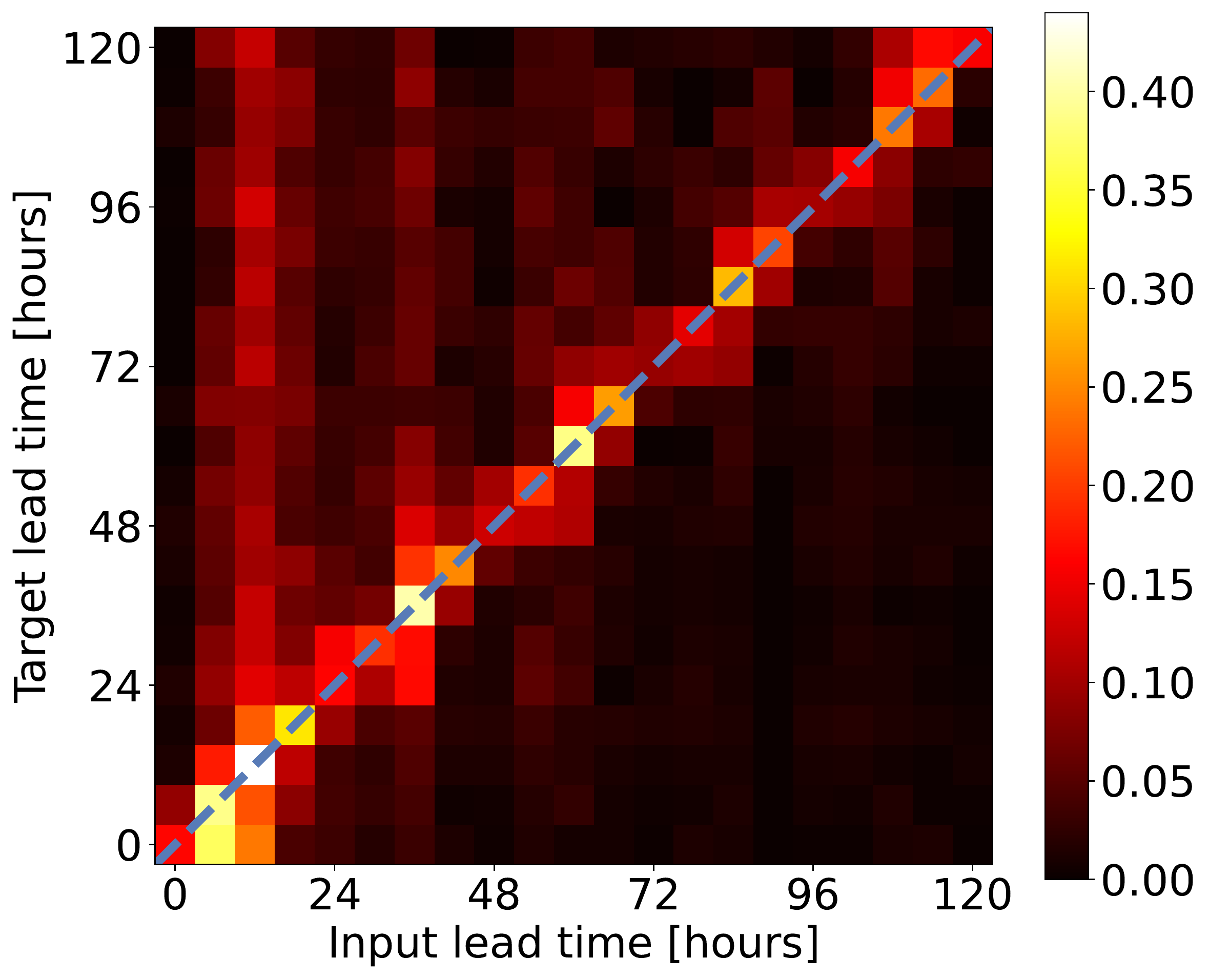}
    \caption{Input forecast lead times ($x$ axis) importance relative to ANET2's post-processed lead time outputs ($y$ axis). Values on the diagonal (blue dashed line) represent the importance of each input lead time to its corresponding post-processed lead time output. Example: the column labeled $24$ contains the importance measures of the forecast issued at that lead time on all lead times in the post-processed output.}
    \label{fig:importance}
\end{figure}

We display the input variable importance relative to a target lead time in Figure \ref{fig:importance}.
Each row signifies a specific lead time in the post-processing output while a column denotes a specific lead time in the ensemble forecast that forms the model input.
In many cases, off diagonal elements of each row exhibit high importance implying that ensemble forecasts for different lead times contribute to the correction for a specific target lead time.
For example, when ANET2 predicts the distribution for the first target lead time (row denoted with zero) it heavily relies on the first three input forecast lead times (the first three columns).
We can also observe that past forecasts exhibit higher importance compared to future forecasts, relative to a specific target lead time.
This is implied by the darker lower triangle of the importance matrix (area under the blue dashed line).
Likewise, we can identify an interesting, periodic trend when observing the columns corresponding to lead times at noon \
We can see that these forecasts non-trivially impact post-processing any target lead time.
We hypothesise that this is due to the daily temperature, which frequently reaches its maximum value at noon (this is due to the six-hour resolution), playing an important role in the amount of heat retained throughout the night (void of all other predictors).
Since the first noon forecast is the most accurate compared to the rest it acts as an estimator, with consecutive noon forecasts decreasing in importance due to forecast errors.

Additionally, looking at the CRPS comparison of ANET1 with other state-of-the-art post-processing techniques evaluate on the EUPPBench dataset (page 17, Figure 3 ion \cite{demaeyer_euppbench_2023}) we can observe that ANET1 exhibits better correction for temperatures at midnight.
The remaining methods suffer from stronger periodic spikes in error when post-processing those lead times.
However, the majority of the implemented methods in \cite{demaeyer_euppbench_2023} (except for reliance calibration) operate on a single lead time.
If we turn our attention again to Figure \ref{fig:importance} and look at each labeled row (corresponding to midnight forecasts), we can see that the importance is more spread out between multiple input lead times.
Therefore, it seems that information important for post-processing forecast at midnight is not solely concentrated at that time but is spread out.
This might explain why our method can better suppress forecast errors at these lead times.

\subsection{Modified derivative estimator}

\begin{figure}[!ht]
    \centering
    \includegraphics[width=0.4\linewidth, valign=t]{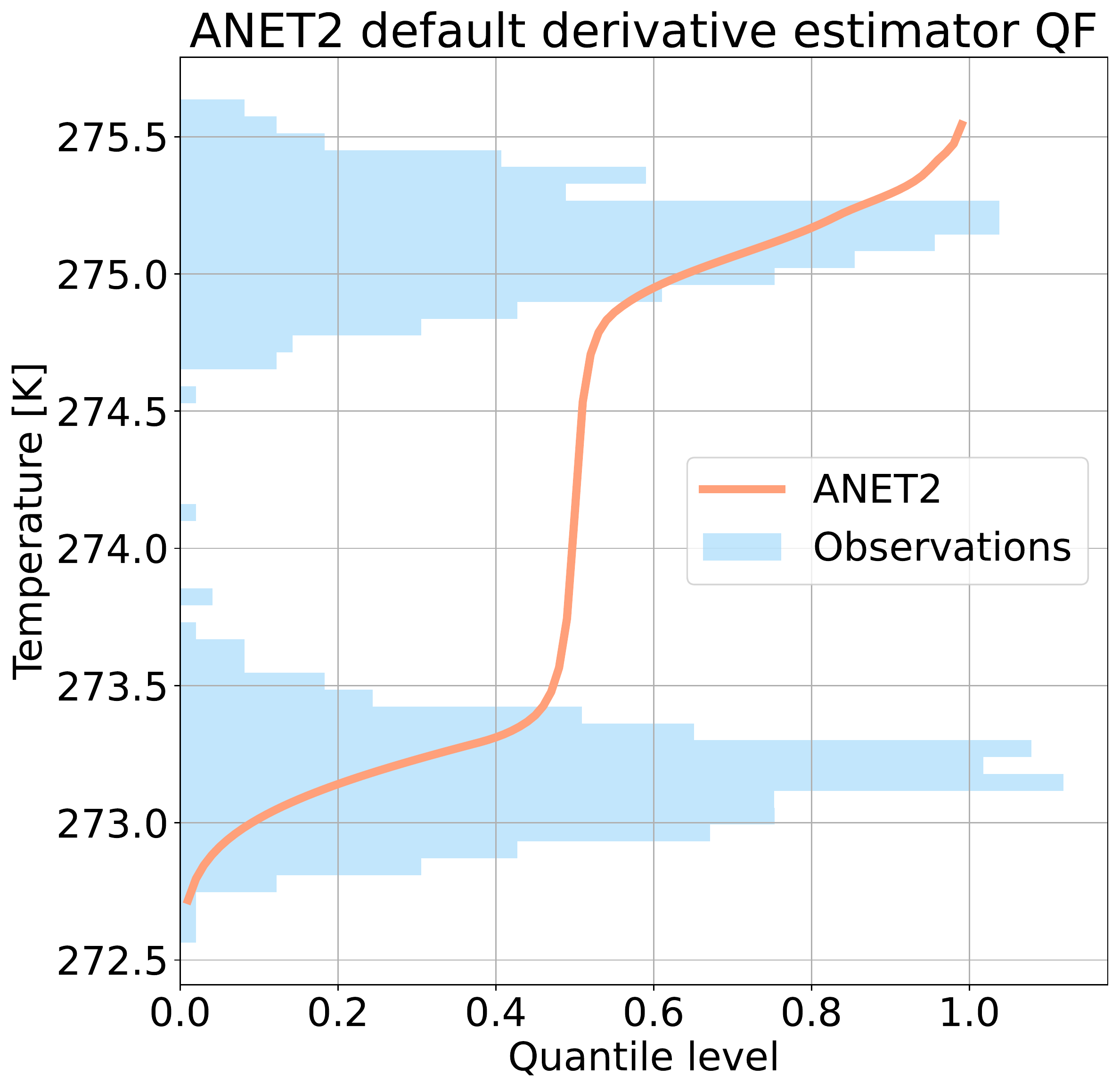}
    \includegraphics[width=0.4\linewidth, valign=t]{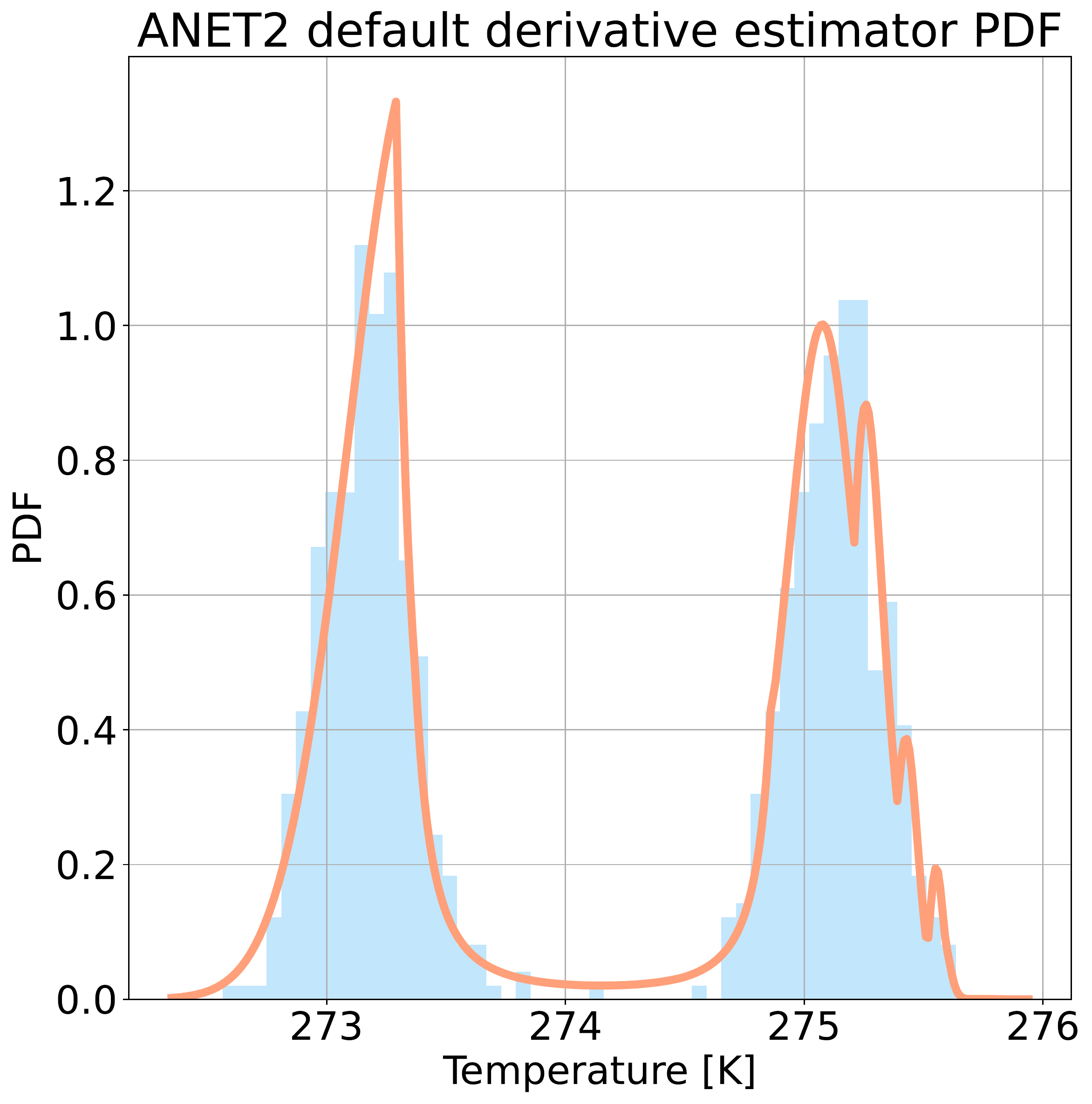}
    \includegraphics[width=0.4\linewidth, valign=t]{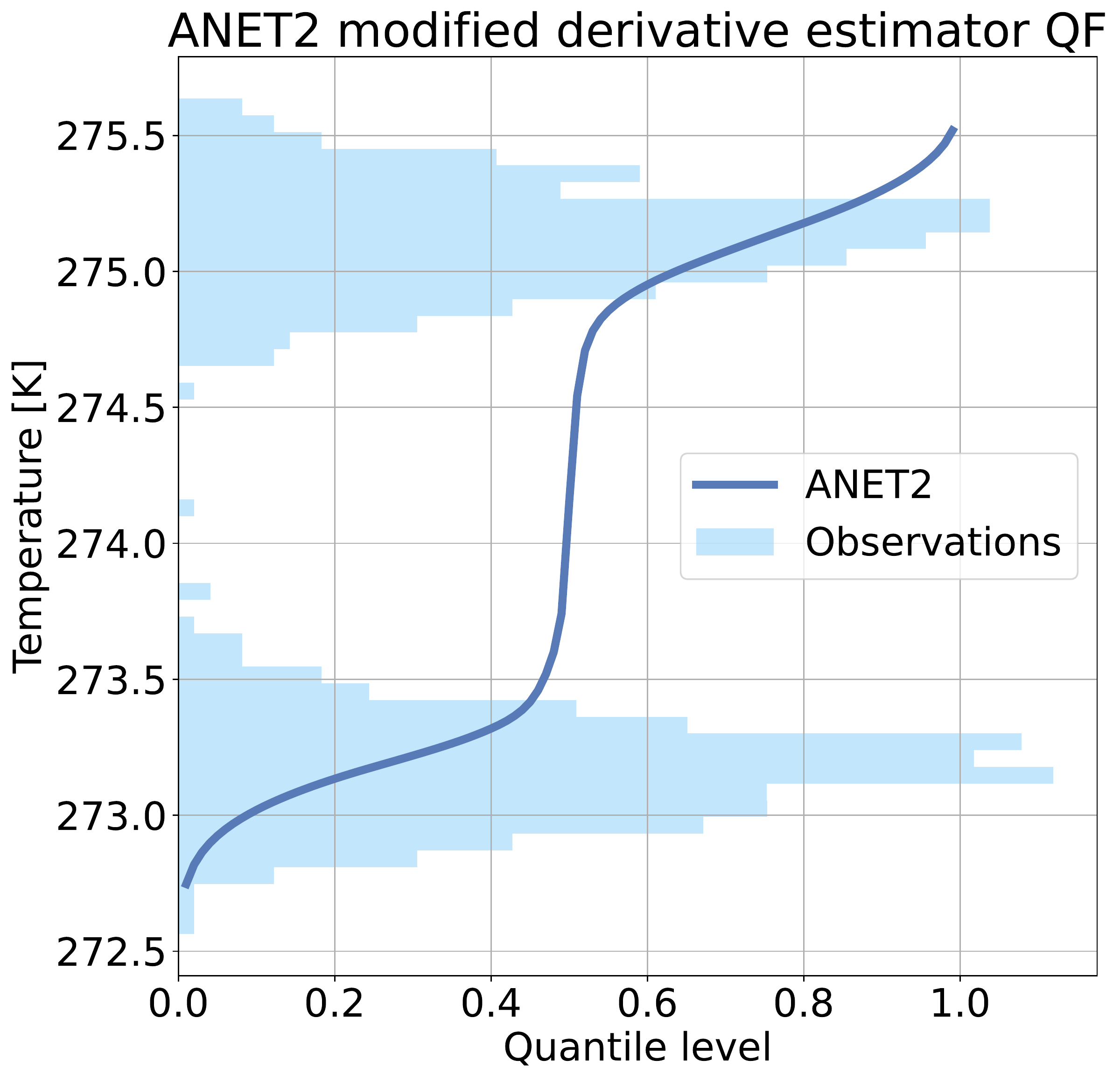}
    \includegraphics[width=0.4\linewidth, valign=t]{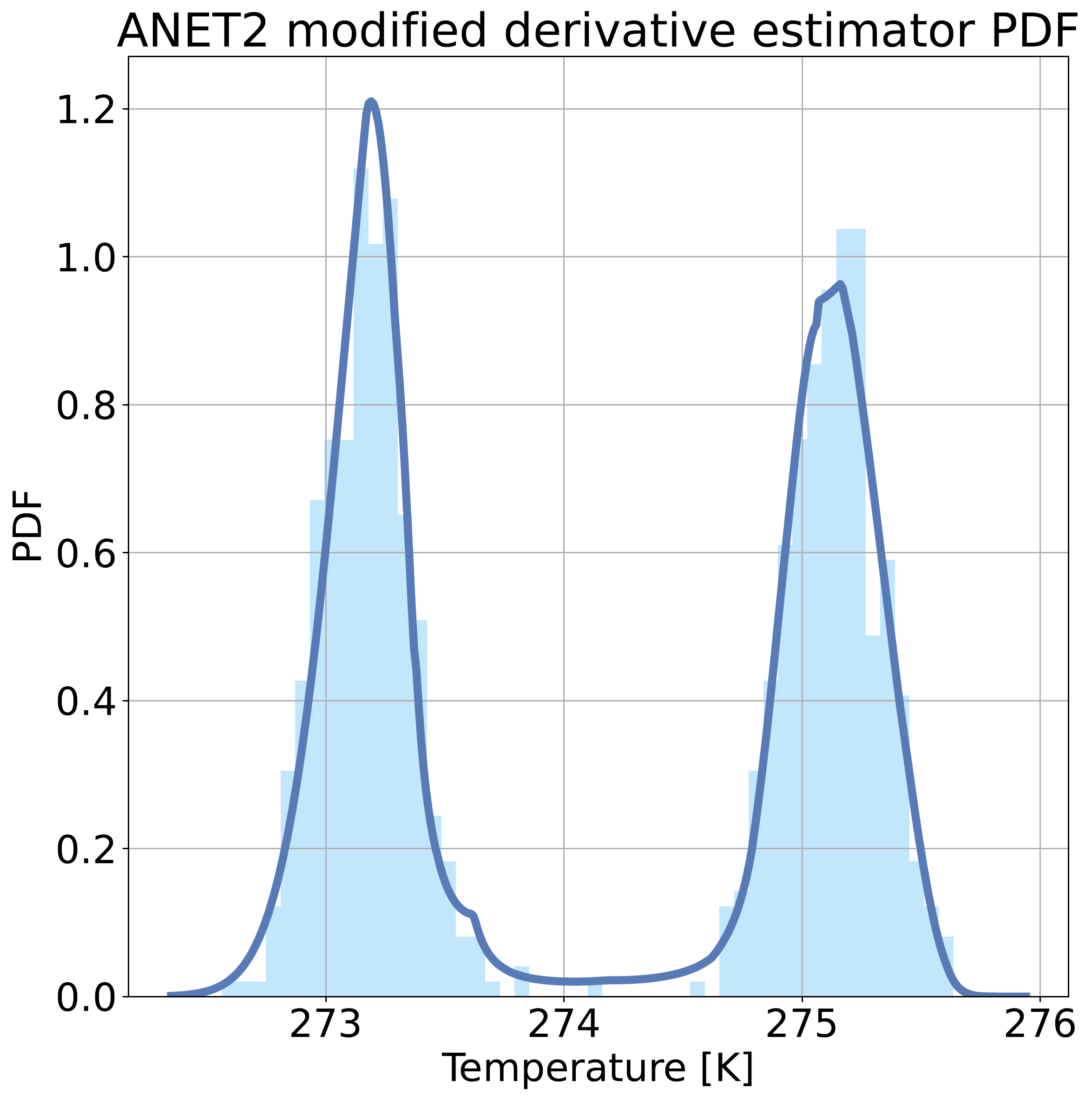}
    \caption{(Top): ANET2 with the default derivative estimator for the normalizing spline flow, as described by \cite{durkan_neural_2019}. (Bottom): ANET2 with the derivative estimation modification based on \cite{gregory_piecewise_1982}. QF denotes the plot containing the quantile function (inverse cumulative distribution function). The ANET2 variant with the modified derivative estimator on the bottom suppresses sharp non-linearities, resulting in more "natural" probability densities and a reduced amount of parameters that the model needs to predict. The observations are generated by sampling two normal distributions with equal variance at different mean temperatures.}
    \label{fig:anet2_derivative}
\end{figure}

In this section we present our modification to the derivative estimator of ANET2's distribution model which is based on normalizing spline flows \cite{durkan_neural_2019}.
The implementation described by \cite{durkan_neural_2019} requires the estimation of three sets of parameters for the spline transformations: spline knots, spline values at the knots, and spline derivatives at the knots.
\cite{durkan_neural_2019} suggest that these parameters be estimated by a neural network.
However, in our testing, this resulted in sharp discontinuities in the probability density of the final estimated distribution, as can be seen in Figure \ref{fig:anet2_derivative}.
This is because the values of the derivatives are selected independently of the knots and values, resulting in potential discontinuities.
\cite{durkan_neural_2019} state that this induces multi-modality into the density.
However, this might hamper future efforts for determining the most likely predicted weather outcome as sharp jumps in the density could introduce noise around the modes.
To produce smoother densities, we modified the derivative estimation procedure such that it now entirely depends on the spline knots and values.
This modified derivative estimator was described by \cite{gregory_piecewise_1982} and offers certain spline continuity guarantees.
The resulting probability density is much smoother (right panel in Figure \ref{fig:anet2_derivative}) and in our tests, the modified distribution estimator performs equally well relative to the default derivative estimator in terms of predictive power in our test scenario.
Another positive attribute of the modified derivative estimation approach is the fact that it reduced the required number of parameters our neural network has to estimate by a third.
This is because in our case the derivatives are estimated from the existing spline knots and values, while the default implementation required the explicit estimation of all three sets of parameters by the neural network.
For example, ANET2 issues a probabilistic forecast for each lead time.
If we use a normalizing spline flow with $s$ consecutive spline transformations, each containing $k$ knots, this results in $21 \times (s \times k \times 3)$ parameters for the default derivative estimation implementation ($21$ for each lead time, times $s$ for each spline, times $k$ for the number of knots, times $3$ for each set of parameters - knots, values, derivatives).
Contrary to that, the modified derivative approach requires $21 \times (s \times k \times 2)$ parameters, as the derivatives are estimated from the remaining parameters.

\subsection{ANET2 drawbacks}

The enhanced distributional flexibility of ANET2 comes with two drawbacks: an increased number of parameters in the distribution model, and increased execution time.
The execution time and distribution model parameter count comparison is displayed in Table \ref{tbl:anet2drw}.
We can see that ANET2 is the slowest of all three variants, exhibiting a roughly five times slower inference time for the $D_{51}$ dataset compared to ANET2$_\text{BERN}$.
ANET2$_\text{NORM}$ has the lowest execution speed, followed by ANET2$_\text{BERN}$.
This difference in time is mainly due to the sequential nature of the normalizing spline flow: input data has to first pass through a cascade of spline transformations before the density can be estimated.
Additionally, when a data sample passes through a spline transformation it has to locate the appropriate spline bin or interval which incurs additional computational burdens of searching for that bin (\cite{durkan_neural_2019} suggest the use of bisection to speed up this search, however, we do not implement this procedure).
The execution times are still small in an absolute sense, as all methods require less than a minute to post-process the entire $D_{51}$ test dataset, which contains $167170$ forecasts.
Nevertheless, we must keep this in mind as these computational burdens might compound in a more complex setting with multiple forecast variables in a spatial context.

We can also see that ANET2 requires the highest number of parameters for determining the distribution model.
Although ANET2 outperforms ANET2$_\text{BERN}$ in all forecast evaluation metrics, we have reached the area of diminishing returns.
ANET2 requires three times as many parameters compared to ANET2$_\text{BERN}$ to produce these improvements.

Still, even in the face of these drawbacks, we believe that ANET2 is the most compelling of the three compared approaches in the EUPPBench setting due to its superior distribution estimation performance, and exact probability distribution estimation with no potential quantile crossing.
Perhaps these would require attention in a more complex forecast post-processing scenario, however, this requires additional research and testing.

\begin{table}[ht]
\begin{center}
\begin{tabularx}{0.835\textwidth}{X | X | X }
  & Inference time for $D_{51}$ test dataset & Number of parameters in distribution model \\
\hline
ANET2$_{\text{NORM}}$ & \textbf{2.87} seconds & \textbf{42} \\
ANET2$_{\text{BERN}}$ & 7.14 seconds & 273 \\
ANET2                 & 36.08 seconds & 840 \\
\end{tabularx}
\end{center}
\caption{The inference execution times and the number of distributional model parameters required by each ANET2 variant. Values in bold are better. The inference times are produced with an Nvidia Quadro P400 graphics card on a system with an Intel i7 9700K processor with $32$ gigabytes of memory.}
\label{tbl:anet2drw}
\end{table}

\section{Conclusion}

In this work we introduced the ANET2 forecast post-processing model to combat two main issues plaguing current state-of-the-art techniques: the lack of expressive power in the parameter estimation and distribution models.
We addressed the first by developing a novel neural network architecture and training procedure, processing the entire forecast lead time at once.
By implementing a modified normalizing flow approach we tackled the second issue, which resulted in the creation of a flexible, mathematically exact distribution model.
We demonstrated the effectiveness of ANET2 against the current state-of-the-art method ANET1, showing that ANET2 outperforms ANET1 in all tested metrics.

This is not to say that ANET2 is without flaws.
From all the compared methods, ANET2 is the slowest in terms of inference time (about $10$ times slower compared to ANET2$_\text{NORM}$, requiring $36$ seconds for the processing of the D$_{51}$ dataset).
Likewise, the distribution model described by the normalizing flow requires the most parameters out of all evaluated models.
While this is not an issue in the context of the initial EUPPBench experiment, these drawbacks have to be kept in mind if one wishes to scale ANET2 to a spatial domain with more than one weather variable to be post-processed at the same time.

Still, we consider ANET2 to be a significant contribution to the field of forecast post-processing. 
To the best of our knowledge, ANET2 is the first method that utilizes normalizing flows to construct weather probability distributions based on weather forecasts. 
When combined with the novel neural network parameter estimation model, ANET2 achieves state-of-the-art post-processing results. 
Our intention in introducing this new tool for weather forecast post-processing is to empower national weather forecast providers to generate higher quality probabilistic models with increased confidence.

\subsection{Data Availability}

The data we used in this study is part of the EUPPBench dataset.
However, a subset of the dataset pertaining to the Swiss station data is not freely available.
For more information about the dataset please refer to \cite{demaeyer_euppbench_2023}. 
The EUPPBench dataset is available at \cite{demaeyer_jonathan_2022_7429236}.

\subsection{Code Availability}

All ANET variants are open source and available on GitHub, with ANET1 accessible at \cite{noauthor_eupp-benchmarkessd-anet_nodate}, and ANET2 at \cite{noauthor_petermlakaranet2_nodate}. 

%%%%%%%%%%%%%%%%%%%%%%%%%%%%%%%%%%%%%%%%%%%%%%%

\section*{Acknowledgments}

We would like to thank EUMETNET for providing the necessary resources, enabling the creation of the EUPPBench benchmark and to our colleagues behind the benchmark, whose dedicated work lead to its realization.
\\
This work was supported by the Slovenian Research Agency (ARRS) research core funding P2-0209 (Jana Faganeli Pucer).

\bibliographystyle{splncs04}
\bibliography{main}

\begin{thebibliography}{10}
\providecommand{\url}[1]{\texttt{#1}}
\providecommand{\urlprefix}{URL }
\providecommand{\doi}[1]{https://doi.org/#1}

\bibitem{ambrogioni_kernel_2017}
Ambrogioni, L., Gü{\c c}lü, U., Gerven, M.A.J.v., Maris, E.: The {Kernel}
  {Mixture} {Network}: {A} {Nonparametric} {Method} for {Conditional} {Density}
  {Estimation} of {Continuous} {Random} {Variables} (2017), \_eprint:
  1705.07111

\bibitem{noauthor_eupp-benchmarkessd-anet_nodate}
benchmark/{ESSD} {ANET}, E.: {ANET1} (2023-03-28),
  \url{https://github.com/EUPP-benchmark/ESSD-ANET}

\bibitem{bauer_quiet_2015}
Bauer, P., Thorpe, A., Brunet, G.: The quiet revolution of numerical weather
  prediction. Nature  \textbf{525}(7567),  47--55 (Sep 2015).
  \doi{10.1038/nature14956}, \url{https://doi.org/10.1038/nature14956}

\bibitem{bremnes_ensemble_2020}
Bremnes, J.B.: Ensemble postprocessing using quantile function regression based
  on neural networks and {Bernstein} polynomials. Monthly Weather Review
  \textbf{148}(1),  403--414 (2020), publisher: American Meteorological Society

\bibitem{chapman_probabilistic_2022}
Chapman, W.E., Delle~Monache, L., Alessandrini, S., Subramanian, A.C., Ralph,
  F.M., Xie, S.P., Lerch, S., Hayatbini, N.: Probabilistic predictions from
  deterministic atmospheric river forecasts with deep learning. Monthly Weather
  Review  \textbf{150}(1),  215--234 (2022)

\bibitem{demaeyer_euppbench_2023}
Demaeyer, J., Bhend, J., Lerch, S., Primo, C., Van~Schaeybroeck, B., Atencia,
  A., Ben~Bouallègue, Z., Chen, J., Dabernig, M., Evans, G., Faganeli~Pucer,
  J., Hooper, B., Horat, N., Jobst, D., Merše, J., Mlakar, P., Möller, A.,
  Mestre, O., Taillardat, M., Vannitsem, S.: The {EUPPBench} postprocessing
  benchmark dataset v1.0. Earth System Science Data Discussions  \textbf{2023},
   1--25 (2023). \doi{10.5194/essd-2022-465},
  \url{https://essd.copernicus.org/preprints/essd-2022-465/}

\bibitem{demaeyer_jonathan_2022_7429236}
Demaeyer, J.: {EUPPBench postprocessing benchmark dataset - gridded data - Part
  I} (Dec 2022). \doi{10.5281/zenodo.7429236},
  \url{https://doi.org/10.5281/zenodo.7429236}

\bibitem{dinh_density_2016}
Dinh, L., Sohl-Dickstein, J., Bengio, S.: Density estimation using real nvp.
  arXiv preprint arXiv:1605.08803  (2016)

\bibitem{dupuy_arpege_2021}
Dupuy, F., Mestre, O., Serrurier, M., Burdá, V.K., Zamo, M.,
  Cabrera-Gutiérrez, N.C., Bakkay, M.C., Jouhaud, J.C., Mader, M.A., Oller,
  G.: {ARPEGE} {Cloud} {Cover} {Forecast} {Postprocessing} with {Convolutional}
  {Neural} {Network}. Weather and Forecasting  \textbf{36}(2),  567 -- 586
  (2021). \doi{https://doi.org/10.1175/WAF-D-20-0093.1},
  \url{https://journals.ametsoc.org/view/journals/wefo/36/2/WAF-D-20-0093.1.xml},
  place: Boston MA, USA Publisher: American Meteorological Society

\bibitem{durkan_neural_2019}
Durkan, C., Bekasov, A., Murray, I., Papamakarios, G.: Neural spline flows.
  Advances in neural information processing systems  \textbf{32} (2019)

\bibitem{ecmwf}
{ECMWF}: \url{https://www.ecmwf.int/en/forecasts} (2022)

\bibitem{fisher_all_2019}
Fisher, A., Rudin, C., Dominici, F.: All {Models} are {Wrong}, but {Many} are
  {Useful}: {Learning} a {Variable}'s {Importance} by {Studying} an {Entire}
  {Class} of {Prediction} {Models} {Simultaneously}. Journal of Machine
  Learning Research  \textbf{20}(177),  1--81 (2019),
  \url{http://jmlr.org/papers/v20/18-760.html}

\bibitem{gneiting_probabilistic_2007}
Gneiting, T., Balabdaoui, F., Raftery, A.E.: Probabilistic forecasts,
  calibration and sharpness. Journal of the Royal Statistical Society. Series
  B: Statistical Methodology  \textbf{69}(2) (2007).
  \doi{10.1111/j.1467-9868.2007.00587.x}

\bibitem{gneiting_probabilistic_2023}
Gneiting, T., Lerch, S., Schulz, B.: Probabilistic solar forecasting:
  {Benchmarks}, post-processing, verification. Solar Energy  \textbf{252},
  72--80 (2023). \doi{https://doi.org/10.1016/j.solener.2022.12.054},
  \url{https://www.sciencedirect.com/science/article/pii/S0038092X22009343}

\bibitem{gneiting_calibrated_2005}
Gneiting, T., Raftery, A.E., Westveld, A.H., Goldman, T.: Calibrated
  probabilistic forecasting using ensemble model output statistics and minimum
  {CRPS} estimation. Monthly Weather Review  \textbf{133}(5) (2005).
  \doi{10.1175/MWR2904.1}

\bibitem{gregory_piecewise_1982}
Gregory, J.A., Delbourgo, R.: Piecewise {Rational} {Quadratic} {Interpolation}
  to {Monotonic} {Data}. IMA Journal of Numerical Analysis  \textbf{2}(2),
  123--130 (Apr 1982). \doi{10.1093/imanum/2.2.123},
  \url{https://doi.org/10.1093/imanum/2.2.123}, \_eprint:
  https://academic.oup.com/imajna/article-pdf/2/2/123/2267745/2-2-123.pdf

\bibitem{gronquist_deep_2021}
Grönquist, P., Yao, C., Ben-Nun, T., Dryden, N., Dueben, P., Li, S., Hoefler,
  T.: Deep learning for post-processing ensemble weather forecasts.
  Philosophical Transactions of the Royal Society A: Mathematical, Physical and
  Engineering Sciences  \textbf{379}(2194),  20200092 (Feb 2021).
  \doi{10.1098/rsta.2020.0092}, \url{https://doi.org/10.1098/rsta.2020.0092},
  publisher: Royal Society

\bibitem{hakim_weather_2017}
Hakim, G., Patoux, J.: Weather: {A} {Concise} {Introduction}. Cambridge
  University Press (2017), \url{https://books.google.si/books?id=pqXoAQAACAAJ}

\bibitem{hamill_interpretation_2001}
Hamill, T.M.: Interpretation of {Rank} {Histograms} for {Verifying} {Ensemble}
  {Forecasts}. Monthly Weather Review  \textbf{129}(3),  550 -- 560 (2001).
  \doi{10.1175/1520-0493(2001)129<0550:IORHFV>2.0.CO;2},
  \url{https://journals.ametsoc.org/view/journals/mwre/129/3/1520-0493_2001_129_0550_iorhfv_2.0.co_2.xml},
  place: Boston MA, USA Publisher: American Meteorological Society

\bibitem{hendrycks_gaussian_2020}
Hendrycks, D., Gimpel, K.: Gaussian {Error} {Linear} {Units} ({GELUs}). arXiv
  preprint arXiv:1606.08415  (2023)

\bibitem{hewson_low-cost_2021}
Hewson, T.D., Pillosu, F.M.: A low-cost post-processing technique improves
  weather forecasts around the world. Communications Earth \& Environment
  \textbf{2}(1), ~132 (2021), publisher: Nature Publishing Group UK London

\bibitem{jaini_tails_2020}
Jaini, P., Kobyzev, I., Yu, Y., Brubaker, M.: Tails of {Lipschitz} {Triangular}
  {Flows}. In: III, H.D., Singh, A. (eds.) Proceedings of the 37th
  {International} {Conference} on {Machine} {Learning}. Proceedings of
  {Machine} {Learning} {Research}, vol.~119, pp. 4673--4681. PMLR (Jul 2020),
  \url{https://proceedings.mlr.press/v119/jaini20a.html}

\bibitem{kingma_adam_2017}
Kingma, D.P., Ba, J.: Adam: {A} {Method} for {Stochastic} {Optimization}. arXiv
  preprint arXiv:1412.6980  (2017)

\bibitem{kirkwood_framework_2021}
Kirkwood, C., Economou, T., Odbert, H., Pugeault, N.: A framework for
  probabilistic weather forecast post-processing across models and lead times
  using machine learning. Philosophical Transactions of the Royal Society A
  \textbf{379}(2194),  20200099 (2021), publisher: The Royal Society Publishing

\bibitem{kobyzev_normalizing_2020}
Kobyzev, I., Prince, S.J., Brubaker, M.A.: Normalizing flows: {An} introduction
  and review of current methods. IEEE transactions on pattern analysis and
  machine intelligence  \textbf{43}(11),  3964--3979 (2020), publisher: IEEE

\bibitem{lerch_convolutional_2022}
Lerch, S., Polsterer, K.L.: Convolutional autoencoders for spatially-informed
  ensemble post-processing. arXiv preprint arXiv:2204.05102  (2022)

\bibitem{mcgovern_using_2017}
McGovern, A., Elmore, K.L., Gagne, D.J., Haupt, S.E., Karstens, C.D.,
  Lagerquist, R., Smith, T., Williams, J.K.: Using {Artificial} {Intelligence}
  to {Improve} {Real}-{Time} {Decision}-{Making} for {High}-{Impact} {Weather}.
  Bulletin of the American Meteorological Society  \textbf{98}(10),  2073 --
  2090 (2017). \doi{https://doi.org/10.1175/BAMS-D-16-0123.1},
  \url{https://journals.ametsoc.org/view/journals/bams/98/10/bams-d-16-0123.1.xml},
  place: Boston MA, USA Publisher: American Meteorological Society

\bibitem{noauthor_petermlakaranet2_nodate}
Mlakar, P.: {ANET2} (2023-03-28), \url{https://github.com/petermlakar/ANET2}

\bibitem{moller_vine_2018}
Möller, A., Spazzini, L., Kraus, D., Nagler, T., Czado, C.: Vine copula based
  post-processing of ensemble forecasts for temperature. arXiv preprint
  arXiv:1811.02255  (2018)

\bibitem{phipps_evaluating_2022}
Phipps, K., Lerch, S., Andersson, M., Mikut, R., Hagenmeyer, V., Ludwig, N.:
  Evaluating ensemble post-processing for wind power forecasts. Wind Energy
  \textbf{25}(8),  1379--1405 (2022), publisher: Wiley Online Library

\bibitem{noauthor_pytorch_nodate}
{PyTorch}: \url{https://pytorch.org} (2023-03-28)

\bibitem{noauthor_softplus_nodate}
Pytorch: Softplus - documentation.
  \url{https://pytorch.org/docs/stable/generated/torch.nn.Softplus.html}
  (2023-03-28)

\bibitem{raftery_using_2005}
Raftery, A.E., Gneiting, T., Balabdaoui, F., Polakowski, M.: Using {Bayesian}
  model averaging to calibrate forecast ensembles. Monthly Weather Review
  \textbf{133}(5) (2005). \doi{10.1175/MWR2906.1}

\bibitem{rasp_neural_2018}
Rasp, S., Lerch, S.: Neural networks for postprocessing ensemble weather
  forecasts. Monthly Weather Review  \textbf{146}(11),  3885--3900 (2018),
  publisher: American Meteorological Society

\bibitem{scheuerer_using_2020}
Scheuerer, M., Switanek, M.B., Worsnop, R.P., Hamill, T.M.: Using artificial
  neural networks for generating probabilistic subseasonal precipitation
  forecasts over {California}. Monthly Weather Review  \textbf{148}(8),
  3489--3506 (2020), publisher: American Meteorological Society

\bibitem{schulz_post-processing_2021}
Schulz, B., Ayari, M.E., Lerch, S., Baran, S.: Post-processing numerical
  weather prediction ensembles for probabilistic solar irradiance forecasting.
  Solar Energy  \textbf{220},  1016--1031 (2021).
  \doi{https://doi.org/10.1016/j.solener.2021.03.023},
  \url{https://www.sciencedirect.com/science/article/pii/S0038092X21002097}

\bibitem{schulz_machine_2022}
Schulz, B., Lerch, S.: Machine {Learning} {Methods} for {Postprocessing}
  {Ensemble} {Forecasts} of {Wind} {Gusts}: {A} {Systematic} {Comparison}.
  Monthly Weather Review  \textbf{150}(1),  235--257 (Jan 2022).
  \doi{10.1175/MWR-D-21-0150.1},
  \url{https://journals.ametsoc.org/view/journals/mwre/150/1/MWR-D-21-0150.1.xml},
  place: Boston MA, USA Publisher: American Meteorological Society

\bibitem{srivastava_dropout_2014}
Srivastava, N., Hinton, G., Krizhevsky, A., Sutskever, I., Salakhutdinov, R.:
  Dropout: {A} {Simple} {Way} to {Prevent} {Neural} {Networks} from
  {Overfitting}. Journal of Machine Learning Research  \textbf{15}(56),
  1929--1958 (2014), \url{http://jmlr.org/papers/v15/srivastava14a.html}

\bibitem{van_schaeybroeck_ensemble_2015}
Van~Schaeybroeck, B., Vannitsem, S.: Ensemble post-processing using
  member-by-member approaches: theoretical aspects. Quarterly Journal of the
  Royal Meteorological Society  \textbf{141}(688),  807--818 (2015), publisher:
  Wiley Online Library

\bibitem{vannitsem_statistical_2021}
Vannitsem, S., Bremnes, J.B., Demaeyer, J., Evans, G.R., Flowerdew, J., Hemri,
  S., Lerch, S., Roberts, N., Theis, S., Atencia, A., Bouallègue, Z.B., Bhend,
  J., Dabernig, M., de~Cruz, L., Hieta, L., Mestre, O., Moret, L., Plenković,
  I.O., Schmeits, M., Taillardat, M., van~den Bergh, J., van Schaeybroeck, B.,
  Whan, K., Ylhaisi, J.: Statistical postprocessing for weather forecasts
  review, challenges, and avenues in a big data world. Bulletin of the American
  Meteorological Society  \textbf{102}(3) (2021).
  \doi{10.1175/BAMS-D-19-0308.1}

\bibitem{veldkamp_statistical_2021}
Veldkamp, S., Whan, K., Dirksen, S., Schmeits, M.: Statistical postprocessing
  of wind speed forecasts using convolutional neural networks. Monthly Weather
  Review  \textbf{149}(4),  1141--1152 (2021)

\end{thebibliography}

\end{document}